\documentclass[10pt,twocolumn,letterpaper]{article}

\usepackage{cvpr}              

\usepackage{graphicx}
\usepackage{amsmath}
\usepackage{amssymb}
\usepackage{booktabs}
\usepackage{color, xcolor}
\usepackage{multirow}
\usepackage{makecell}
\usepackage{comment}
\usepackage[accsupp]{axessibility}  

\usepackage[pagebackref,breaklinks,colorlinks]{hyperref}

\usepackage[capitalize]{cleveref}
\crefname{section}{Sec.}{Secs.}
\Crefname{section}{Section}{Sections}
\Crefname{table}{Table}{Tables}
\crefname{table}{Tab.}{Tabs.}

\def\cvprPaperID{3219} 
\def\confName{CVPR}
\def\confYear{2023}

\begin{document}

\title{ReasonNet: End-to-End Driving with Temporal and Global Reasoning}

\author{%
Hao Shao$^{1}$~~~~~~ Letian Wang$^{2}$~~~~~~ Ruobing Chen$^1$ \\ \vspace{0.5em}Steven L. Waslander$^{2}$~~ Hongsheng Li$^3$~~ Yu Liu$^{1,4}$\thanks{Corresponding author}\\ 
$^1$SenseTime Research ~~~ $^2$University of Toronto ~~~  $^3$CUHK MMLab \\   $^4$Shanghai Artificial Intelligence Laboratory
}

\maketitle

\begin{abstract}
The large-scale deployment of autonomous vehicles is yet to come, and one of the major remaining challenges lies in urban dense traffic scenarios. In such cases, it remains challenging to predict the future evolution of the scene and future behaviors of objects, and to deal with rare adverse events such as the sudden appearance of occluded objects. In this paper, we present ReasonNet, a novel end-to-end driving framework that extensively exploits both temporal and global information of the driving scene. By reasoning on the temporal behavior of objects, our method can effectively process the interactions and relationships among features in different frames. Reasoning about the global information of the scene can also improve overall perception performance and benefit the detection of adverse events, especially the anticipation of potential danger from occluded objects. For comprehensive evaluation on occlusion events, we also release publicly a driving simulation benchmark DriveOcclusionSim consisting of diverse occlusion events. We conduct extensive experiments on multiple CARLA benchmarks, where our model outperforms all prior methods, ranking first on the sensor track of the public CARLA Leaderboard\cite{leaderboard}.
\end{abstract}


\section{Introduction}
Despite significant recent progress in the field of autonomous driving, truly large-scale deployment of autonomous vehicles (AVs) on public roads has yet to be established. The majority of the remaining issues lie in navigating dense urban traffic scenes, where a large number of different dynamic objects (e.g. vehicles, bicycles, pedestrians), complex road geometries and road user interactions are involved. In such circumstances, currently deployed or tested solutions could make incorrect or unexpected decisions , resulting in severe accidents or traffic infractions~\cite{tesla, uber, leaderboard}. Two of the major challenges behind such autonomous incompetence include 1) how to achieve a comprehensive understanding of the driving scene and, more importantly, how to make high-fidelity predictions on the future evolution of the driving scene; 2) how to deal with rare adverse events in long-tail distributions, such as undetected but relevant objects in occluded regions.

\begin{figure}[t]
    \centering
    \includegraphics[width=0.45\textwidth]{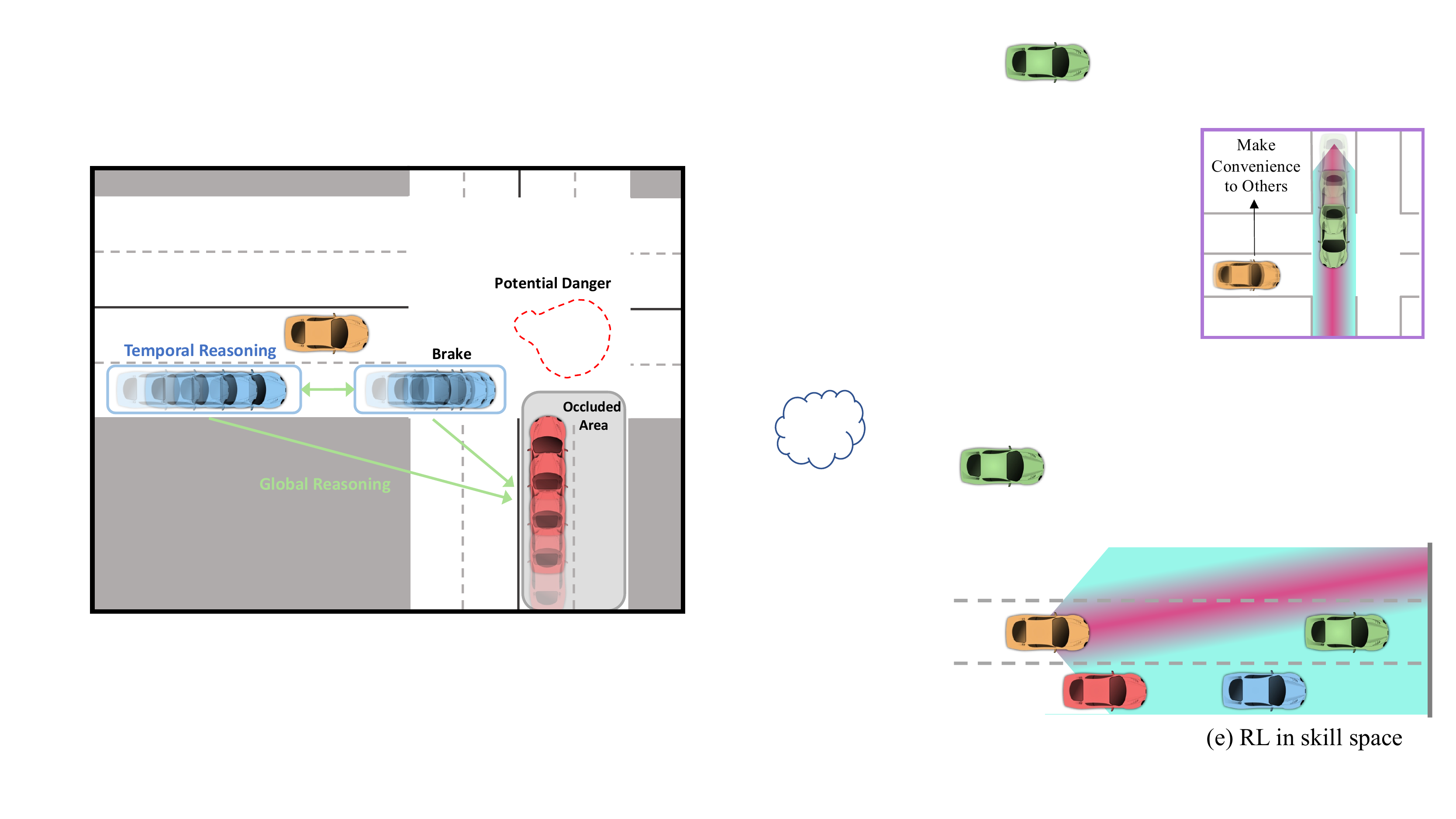}
    \vspace{-0.5em}
    \caption{Temporal reasoning on the historic behaviors of surrounding objects can benefit the prediction of the scene evolution and objects' future behaviors. Global reasoning on the interaction among objects and the environment allows for inference about unobservable space and occluded objects, anticipating potential danger and enhancing perception/driving performance.}
    \label{fig:pipeline}
    \vspace{-1em}
\end{figure}

Comprehensive scene understanding and high-fidelity prediction of how objects in the scene will move in the future are vital for autonomous vehicles to take safe and reliable actions. Toward this end, modularized methods were proposed to decompose the task into three sequential sub-tasks: detection~\cite{li2022bevformer, liu2022bevfusion, lian2022semi, lian2022monojsg, lian2022exploring}, tracking~\cite{zhang2022bytetrack,cao2022observation}, and forecasting~\cite{jia2022hdgt,gu2021densetnt,wang2022transferable,wang2021socially,casas2020implicit,hu2021fiery,wei2021perceive,jia2023towards}. While more interpretability is provided by developing each module independently, these sub-tasks are still regarded as open research questions and errors in each sub-task can propagate and accumulate, leading to unstable overall performance. In contrast, end-to-end driving methods~\cite{shao2022safety, chen2022learning, chen2021learning} have recently emerged as a promising method to solve these subtasks in a monolithic manner directly. However, a high-fidelity future prediction necessitates sufficient \textbf{temporal reasoning} over the historic information of the scene~\cite{feichtenhofer2019slowfast,shao2020temporal}, which is usually only somewhat considered in previous end-to-end driving methods if not completely ignored. For example, \cite{chitta2022transfuser,chen2022learning} only exploited scene information in the current frame, and~\cite{toromanoff2020end} simply concatenated features in historic frames for temporal reasoning. In such cases, the interactions and relationships amongst features in different frames and objects cannot be sufficiently modeled. Thus in this paper, we propose a temporal reasoning module to effectively fuse information from different frames for better driving performance.

On the other hand, rare adverse events in long-tail distributions remain a notoriously challenging issue on the way toward large scale deployment of autonomous vehicles. For example, one such challenge is the difficulty in detecting occluded but relevant objects in the scene. While a large amount of research has focused on improving perception performance~\cite{li2022bevformer, vora2020pointpainting}, the occluded objects essentially lie out of the scope of observable elements, and failure to consider such objects can result in either dangerous or overly cautious driving behavior. Our observation is that, while humans also suffer from similar limitations to autonomous vehicles regarding occluded objects, they are able to reason about these unobservable spaces by exploiting global information of the scene such as road geometry and driving interaction patterns, to anticipate potential danger even under occlusion. For example, when one human driver notices another vehicle braking abruptly, the driver may reason the presence of an occluded object (e.g., a pedestrian) ahead, reminding himself to drive cautiously. Thus, our insight is that, a safe and intelligent autonomous vehicle should also master the \textbf{global reasoning} capability to have a better perception of the scene. In this paper, we propose a transformer-based global reasoning module to sufficiently fuse information of the environment and objects, and  analyze  their interactions for better scene understanding. Such global reasoning capability not only benefits interaction modeling with occluded objects, but also improves overall perception performance. Examples of such performance gains include better traffic light status identification by reasoning over other vehicles' actions and more accurate future trajectory forecasting by reasoning over interactions among objects.
Besides, considering the fact that the occlusion events lie in the long-tail distribution and have been rare in currently available datasets, we also construct a Driving in Occlusion Simulation benchmark (DOS) consisting of 4 occlusion scenarios, each with 25 cases, as a comprehensive occlusion event evaluation benchmark in the field of end-to-end autonomous driving.

In this paper, we propose a novel end-to-end driving framework named temporal and global reasoning network (ReasonNet), which provides enhanced reasoning on the temporal evolution and the global information of the scene, for better perception performance and driving quality. Our contributions are three-fold:

\begin{itemize}
    \item We propose a novel temporal and global reasoning Network (ReasonNet) to enhance historic scene reasoning for high-fidelity prediction of the scene's future evolution and improve global contextual perception performance even under occlusion.
    \item We present a new benchmark called \textbf{D}riving in \textbf{O}cclusion \textbf{S}imulation benchmark (DOS), which consists of diverse occlusion scenarios in urban driving for systematic evaluation in occlusion events, and make the benchmark publicly available.
    \item We experimentally validate our method on multiple benchmarks with complex and adversarial urban scenarios. Our model ranks first on the sensor track of the CARLA autonomous driving leaderboard.
\end{itemize}

\section{Related work}
\label{sec: related work}
\noindent\textbf{End-to-end Autonomous Driving}
End-to-end autonomous driving in urban scenarios has become more studied recently thanks to the CARLA simulator and leaderboard~\cite{dosovitskiy2017carla}. 
Recent works mainly consist of reinforcement learning (RL) and imitation learning (IL) methods. 
The reinforcement Learning methods train the agents by constantly interacting with simulated environments and learning from these experiences. 
Latent DRL~\cite{toromanoff2020end} first trains an embedding space as a latent representation of the environment observation, and then conducts reinforcement learning with the latent observation. 
Roach~\cite{zhang2021end} utilizes an RL agent with privileged information of the environment to distill a model only with regular information (e.g. sensor) as the final agent.
WOR~\cite{chen2021learning} builds a model-based RL agent along with the world model and reward model. The final agent is distilled from the expert knowledge acquired from these pretrained models.
Imitation learning methods aim at learning from an expert agent to bypass interacting with the environment. Early IL methods include CIL~\cite{codevilla2018end} and CILRS~\cite{codevilla2019exploring}, which apply a conditional architecture with different network branches for different navigation commands. 
LBC~\cite{chen2020learning} first trains an imitation learning agent with privileged information, which is then distilled into a model using sensor data.
Transfuser~\cite{prakash2021multi, chitta2022transfuser} designs a multi-modal transformer to fuse information between the front camera image and LiDAR data.
LAV~\cite{chen2022learning} exploits data of not only the ego vehicle but also surrounding vehicles for data augmentation by learning a viewpoint-invariant spatial intermediate representation. 
TCP~\cite{wu2022trajectory} proposes a network with two branches which generates the control signal and waypoints respectively. An adaptive ensemble is applied to fuse the two output signals.
InterFuser~\cite{shao2022safety} uses a transformer to fuse and process multimodal multi-view sensors for comprehensive scene understanding.

\noindent\textbf{Attention for Autonomous Driving}
The attention mechanism has been demonstrated to be a powerful module in many areas of deep learning, including the context of driving. The classic attention-based Transformer architecture~\cite{vaswani2017attention} was originally established in Natural Language Processing. Transformer (VIT) was then applied in Computer Vision (vision Transformer, VIT~\cite{dosovitskiy2020image, qian2021blending}) and attains excellent performance on Imagenet classification.
Later generations move on to generalize the attention mechanism to the driving domain, including motion forecasting~\cite{li2020end,gao2020vectornet,wang2021hierarchical}, driver attention prediction~\cite{gou2022driver, kim2019grounding} and object tracking~\cite{sun2020transtrack, meinhardt2022trackformer}. In the field of end-to-end autonomous driving, TransFuser~\cite{prakash2021multi, chitta2022transfuser} exploits several transformer modules for the fusion of data from the front view camera and LiDAR.  NEAT~\cite{chitta2021neat} uses intermediate attention maps to iteratively compress 2D image features into a compact bird-eye-view (BEV) representation for driving.  InterFuser~\cite{shao2022safety} utilizes a transformer encoder and decoder to fuse information and decode the feature into interpretable embeddings. 

\noindent\textbf{Multi-task Learning} Our end-to-end driving framework adopts multi-task learning, with a joint objective of object detection, occupancy forecasting, traffic sign prediction and waypoint prediction. MotionNet~\cite{wu2020motionnet} proposes a spatio-temporal pyramid network  for joint perception and motion prediction based on BEV maps. PnPNet~\cite{liang2020pnpnet} proposes a new object trajectory representation and multi-object tracker to handle occlusion and false positives. IntentNet~\cite{casas2018intentnet} predicts the high-level intentions of each agent from semantic HD maps building. ST-P3~\cite{hu2022st} proposes an egocentric-aligned accumulation technique to preserve geometry information in 3D space and utilize a dual pathway modeling to consider past motion variations.

\section{Method}
\label{sec: method}
\begin{figure*}[t]
    \centering
    \includegraphics[width=0.925\textwidth]{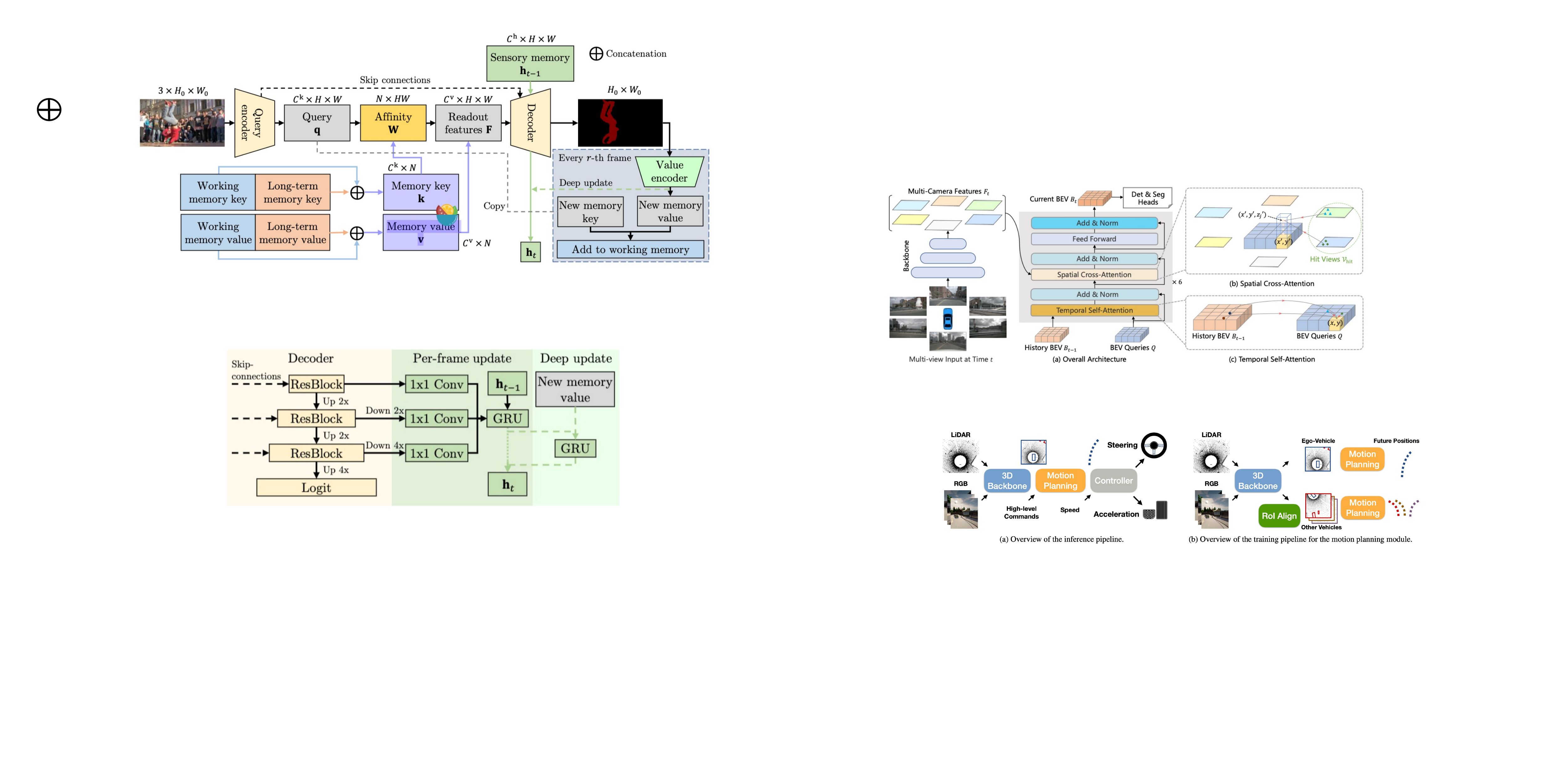}
    \vspace{-0.5em}
    \caption{The proposed ReasonNet consists of three modules: 1) the perception module fuses  different sensor data to generate the BEV feature, traffic sign feature, and waypoints in the early stage of our framework; 2) the temporal reasoning module processes current and historic features and maintains a memory bank to store historic features; 3) the global reasoning module models the interaction and relationship among objects and the environment to detect adverse events (e.g. occlusion) and improve overall perception performance.}
    \label{fig:pipeline}
    \vspace{-1em}
\end{figure*}

We aim at learning a driving policy $\pi$ that generates raw control commands by taking multi-view multi-modal sensor readings, vehicle measurements, and navigation commands as inputs. As shown in Figure~\ref{fig:pipeline}, the proposed ReasonNet consists of three parts: 1) a perception module that extracts bird's-eye-view (BEV) features from LiDAR and RGB data; 2) a temporal reasoning module that processes temporal information and maintains a memory bank storing historic features; 3) a global reasoning module that captures the interaction/relationship amongst objects and the environment, to detect adverse events (e.g. occlusion) and improve overall perception performance. This section will introduce these modules in detail.

\subsection{Perception Module}
\label{sec:perception}

The perception module is responsible for processing and fusing different sensor data at the early stage of our framework, based on which temporal and global reasoning can be conducted by later modules. Specifically, five sensors are utilized: four RGB cameras (left, front, right and rear) $I_{\text{rgb}} = I_{0, 1, 2, 3}$ and one LiDAR sensor $I_{\text{lidar}} = I_{4}$. Four image inputs are obtained from the four cameras, and an additional focus-view image input is center-cropped from the front image to capture distant traffic lights. Point cloud data is retrieved from the LiDAR sensor. Our perception module includes a 2D backbone to embed image input into keys and values, a 3D backbone to embed LiDAR input into queries, and a BEV decoder that utilizes these keys, values, and queries to obtain features of the bird's-eye view (BEV) map, waypoints, and traffic signs.

\noindent\textbf{Image Input} For each image input of $I_{rgb}$, a 2D CNN backbone ResNet~\cite{he2016deep} is applied to generate a feature map $f_{i}$. Then, we use a convolution layer to map the channels of $f_{i}$ to $C_{v}$ and flatten it to one-dimensional tokens. A sinusoidal positional encoding and learnable sensor embeddings are added to the tokens, so that the following network can distinguish them from different cameras and relative positions. Finally, tokens of different images are passed through a standard transformer encoder with $K_e$ layers. Each layer consists of Multi-Headed Self-Attention~\cite{vaswani2017attention}, MLP blocks and layer normalization~\cite{ba2016layer}. This image fusion operation can contribute to a better perception of global context from multi-view inputs, generating keys and values for the image-LiDAR fusion in BEV decoder.

\noindent\textbf{LiDAR Input} For the LiDAR input, we use  PointPillars~\cite{lang2019pointpillars} as our 3D perception backbone to process points in the ego-vehicle-centered area $x \in [- H_{b}, H-H_{b}]$ and $y \in [-W/2, W/2]$. Specifically, we use a simplified version of PointNet\cite{qi2018frustum} to encode the information of raw LiDAR points. Each pillar includes the points in a 0.25m$\times$0.25m area. The extracted feature map is downsampled to $ C_{v} \times H \times W$ for computation reduction and then serves as BEV queries used in the BEV decoder and memory bank.

\noindent\textbf{Sensor-Fusion BEV Decoder} The BEV decoder follows a standard transformer architecture design with $K_{bev}$ layers to fuse tokens from different sensors.
Tokens from the RGB images are fed as values and keys into the decoder, and tokens from the LiDAR points are fed as the $H \times W$ queries to generate BEV features. In addition, two other kinds of queries for the prediction of traffic signs and waypoints $\mathbf{w}$ are also fed into the decoder. Following InterFuser~\cite{shao2022safety}, we use a 2-layer MLP as the traffic sign classifier to predict the traffic light state and whether there is a stop sign ahead; we then use a single-layer GRU~\cite{cho2014learning} to auto-regressively generate consecutive waypoints $\{\mathbf{w}_{t}\}_{t=1}^{T_{f}}$ conditioned on the goal location of the ego vehicle. $T_{f}$ denotes the number of the predicted time steps. To pretrain the perception module in the first training stage, the generated BEV feature is passed through a one-stage CenterPoint~\cite{yin2021center} to generate the $H \times W \times 7$ BEV map covering an $H$m $\times$ $W$m spatial region, where the seven channels represent object existence probability, offset from grid center, bounding box extent, heading angle and velocity for objects at each grid cell.

\subsection{Temporal Reasoning module}
\label{sec:temporal reasoning}

Compared to existing end-to-end driving methods that only exploit scene information of the current frame~\cite{chitta2022transfuser,chen2022learning} or simply concatenate features of historic frames~\cite{toromanoff2020end}, we propose a temporal reasoning module that can sufficiently store and fuse temporal information to benefit the motion forecasting of traffic participants and the tracking of intermittently occluded objects. As shown in Figure~\ref{fig:pipeline}, our temporal reasoning module includes temporal processing to fuse current and historic features through an attention mechanism, and maintains a memory bank which stores historic short-term and long-term feature keys and values.

\noindent\textbf{Temporal Processing} Considering that information in different historic frames could have different relevance to the current scene, we apply an attention-based memory reading from the historic features. Specifically, for each historic frame $t$ stored in the memory, we first measure its relevance by calculating the normalized similarity $\mathbf{S}$ between the historic-frame feature key $\mathbf{k} \in \mathbb{R}^{C_{k} \times T_h \times H \times W}$\footnote{$C$, $H$, $W$ denotes the channel, height, and width of the feature respectively, $T_h$ denotes the number of the frames stored in the memory bank.} and the current-frame feature query $\mathbf{q}\in \mathbb{R}^{C_{k} \times H \times W}$: 
\begin{equation}
\label{equ:similarity}
S(\mathbf{q}_{h, w}, \mathbf{k}_{t, i, j}) = \frac{(\mathbf{k}_{t, i, j} - \mathbf{q}_{h, w})^{2}}{\sum_{i=0,j=0}^{i=H-1,j=W-1} (\mathbf{k}_{t, i, j} - \mathbf{q}_{h, w})^{2}}  
\end{equation}

We map every query element to a distribution over $H \times W$ memory elements and correspondingly aggregate their values v to obtain the readout feature $\mathbf{M} \in \mathbb{R}^{C_{v} \times T_h \times H \times W}$ for each frame $t$ stored in the memory:

\begin{equation}
\label{equ:readout}
\mathbf{M}_{t,h,w} = \sum_{i=0,j=0}^{i=H-1,j=W-1} \mathbf{v}_{t, i, j} S(\mathbf{q}_{h, w}, \mathbf{k}_{t, i, j})
\end{equation}

The aggregated features from all historic frames are then concatenated with the current-frame feature value to get $\mathbf{M}^{'} \in \mathbb{R}^{C_{v} \times (T_h+1) \times H \times W}$, which is then passed through a GRU to progressively fuse temporal information and get $\mathbf{M}_{fused} \in \mathbb{R}^{C_{v} \times H \times W}$ as the final output of the module. Technically, we take the L2 similarity proposed in STCN\cite{cheng2021rethinking} as the similarity measure function, which is more stable than the dot product\cite{patrick2021keeping}. The current-frame feature query $\mathbf{q}$ is obtained by passing the features from the 3D backbone $\mathbf{F} \in \mathbb{R}^{C_{v} \times H\times W}$ through a query encoder (several convolution layers). The historic-frame feature key $\mathbf{k}$ and value $\mathbf{v}$ are taken from the temporal memory bank.

\noindent\textbf{Memory Bank Maintaining} As above, we have introduced the temporal processing at one single frame. After every $\tau$ frame, the obtained feature key and value at that frame will be used to update the memory bank. Specifically, the current-frame feature query $\mathbf{q}$ is directly copied and fed into the memory bank as the memory key without extra computation. The final output $\mathbf{M}_{fused}$ will first be encoded to a BEV map $\mathbf{M}_p$. The BEV map $\mathbf{M}_p$ will be concatenated with the final output $\mathbf{M}_{fused}$ and passed through a value encoder to obtain the memory value $\mathbf{v}$, which is fed into the memory back. With the above key-value pairs, the memory bank maintains two kinds of buffer:  the short-term and long-term buffer. On the one hand, the new key-value pair will be appended to the short-term buffer, as a high-resolution memory of the scene in the past few seconds for accurate feature matching. Considering the limited GPU memory resources, we limit the buffer size and older key-value pairs will be discarded when the limit number $T_{s}$ is reached. However, when these older features are discarded, the long-term behavior of the traffic participants is missing, which can be crucial for motion forecasting in complex traffic scenarios. Thus on the other hand, inspired by XMem\cite{cheng2022xmem}, the memory bank also maintains a long-term buffer that selectively stores important/representative key-value feature pairs discarded by the short-term buffer. Considering the fact that the objects surrounding the ego vehicle are sparse most of the time\footnote{Based on the data collected from the CARLA simulator, only 7\% of the ego-vehicle-centered BEV map area is occupied by active objects.}, the long-term buffer selectively stores key-value features ($\mathbf{k}$ and $\mathbf{v}$) which meet one of the two criteria: 1) their corresponding location in the BEV map $\mathbf{M}_p$ has a high probability of object existence; 2) their usage frequency is in top-\textit{K} of all candidate key-value features. The usage frequency is defined by its cumulative normalized similarity (Eq. \ref{equ:similarity}).
The features selected by the above criteria are appended to the last frame of memory. And if the last memory frame is full, we will initialize one new frame with the zero vector and set it as the last frame to store new features. When the number of frames reaches the limit $T_{l}$, the obsolete memory will be removed.
Such a compact storing strategy can efficiently track long-term representative features and intermittently occluded objects, while balancing the resources required. 

\subsection{Global Reasoning module}
\label{sec:global reasoning}

Rare adverse events such as occluded objects are a notorious issue for the practical deployment of AVs. Our insight is that humans perceive their surroundings not only through sensors, but also by exploiting global information on the scene to reason over the unobservable spaces. For instance, when a vehicle performs an emergency stop without a clear reason, humans can infer that there is potentially an occluded object ahead of the vehicle and will drive more cautiously. Thus we propose the global reasoning module to capture the interaction and relationship between objects and the environment to detect adverse events (e.g. occlusion) and improve overall perception performance. The module consists of three parts: 1) an object-environment and object-object interaction modeling process; 2) an occupancy decoder to generate the occupancy map; 3) a consistency loss to encourage consistent prediction of waypoints and the occupancy map.

\noindent\textbf{Interaction Modeling} The object-environment and object-object interaction modeling process aims at reasoning about the relationship among objects and the environment. On the one hand, $\mathbf{M}_{fused}$ features whose corresponding location in the BEV map $\mathbf{M}_p$ has a high probability of object existence will be extracted to represent object features. On the other hand, $\mathbf{M}_{fused}$ features will also be downsampled to represent the environment features. All object and environment features are used to construct a graph, which is passed through a graph attention network (GAT)~\cite{velickovic2017graph} for interaction modeling.

\noindent\textbf{Occupancy Decoder} Taking the features outputted by the GAT as keys and values, and the learnable positional embeddings as queries, the occupancy decoder utilizes a transformer decoder with $K_{opy}$ layers to generate: 1) the traffic sign feature, which is then concatenated with the traffic sign feature from the BEV decoder to generate the final traffic sign prediction; 2) the occupancy map feature, which is then applied with convolution operation to generate the occupancy map $\mathbf{O}_{t} \in \mathbb{R}^{T_f \times H \times W}$. At a future time $t$, each cell in the occupancy map contains a value in the range [0,1] representing the probability that the cell is occupied.

\noindent\textbf{Consistency Loss} Currently, our framework predicts the waypoints and the occupancy map independently, which are not necessarily consistent. For example, the waypoints could overlap some obstacles in the occupancy map. Thus we propose a consistency loss to discourage waypoints' crossing the high-probability region of the occupancy map. Further, the consistency loss also encourages generating longer waypoint trajectories for efficient driving. Specifically, the consistency loss aims at minimizing the average object existence probability of the cells that cover the predicted waypoints, and maximizing the average $l_2$ length of the waypoint trajectory $\mathbf{w}$:

\begin{equation}
    \mathcal{L}_{\text{consistency}} = \frac{1}{T_f}\Bigg( \sum_{t=0}^{T_f} \frac{\sum_{i=0}^{N_{c}^{t}} \mathbf{O}_{t, i}}{N_{c}^{t}} - \lambda  \sum_{t=0}^{T_f}\| \mathbf{w}_{t} \|_{1} \Bigg)
\end{equation}
,where $N_c^t$ denotes the number of covered cell at step $t$, $\mathbf{O}_{t, i}$ denotes the object existence probability at cell $i$ at time $t$.

\subsection{Control}
\label{sec:control}
Following~\cite{chen2020learning}, we use two PID controllers for latitudinal and longitudinal control, to track the heading and velocity of predicted waypoints respectively. If a red traffic light or stop sign is detected, the ego-vehicle will brake. Additionally, an emergency stop will also be performed if the ego vehicle's current bounding box crosses the area in the occupancy map that has a high object existence probability or if the future waypoints overlap with objects in the BEV map.

\subsection{Training Setup}
\label{sec:training setup}
The training of our framework consists of two stages. In the first stage, we train the perception module to predict BEV features, traffic sign features, and waypoints. Specifically, the loss of BEV features and traffic sign features is computed with additional prediction heads, which are discarded in the next stage. In the second stage, we freeze the perception module and train the other two modules. Five loss terms are considered: 1) the waypoints loss $\mathcal{L}_{w}$ that minimizes the error between predicted waypoints and expert waypoints; 2) the BEV map loss $\mathcal{L}_{BEV}$ that follows~\cite{yin2021center,chen2022learning} to minimize the current-frame BEV map prediction error; 3) the traffic sign loss $\mathcal{L}_{sign}$ for the traffic regulation prediction; 4) occupancy map loss $\mathcal{L}_{opy}$ that minimizes the occupancy prediction error in a future horizon; 5) the consistency loss $\mathcal{L}_{\text{consistency}}$ which encourages a consistent generation of the waypoints and occupancy map. These loss terms are balanced by corresponding loss weights.

\section{Drive in Occlusion Sim (DOS) Benchmark}
\label{sec:dos}
\begin{figure}[t]
    \centering
    \includegraphics[width=0.45\textwidth]{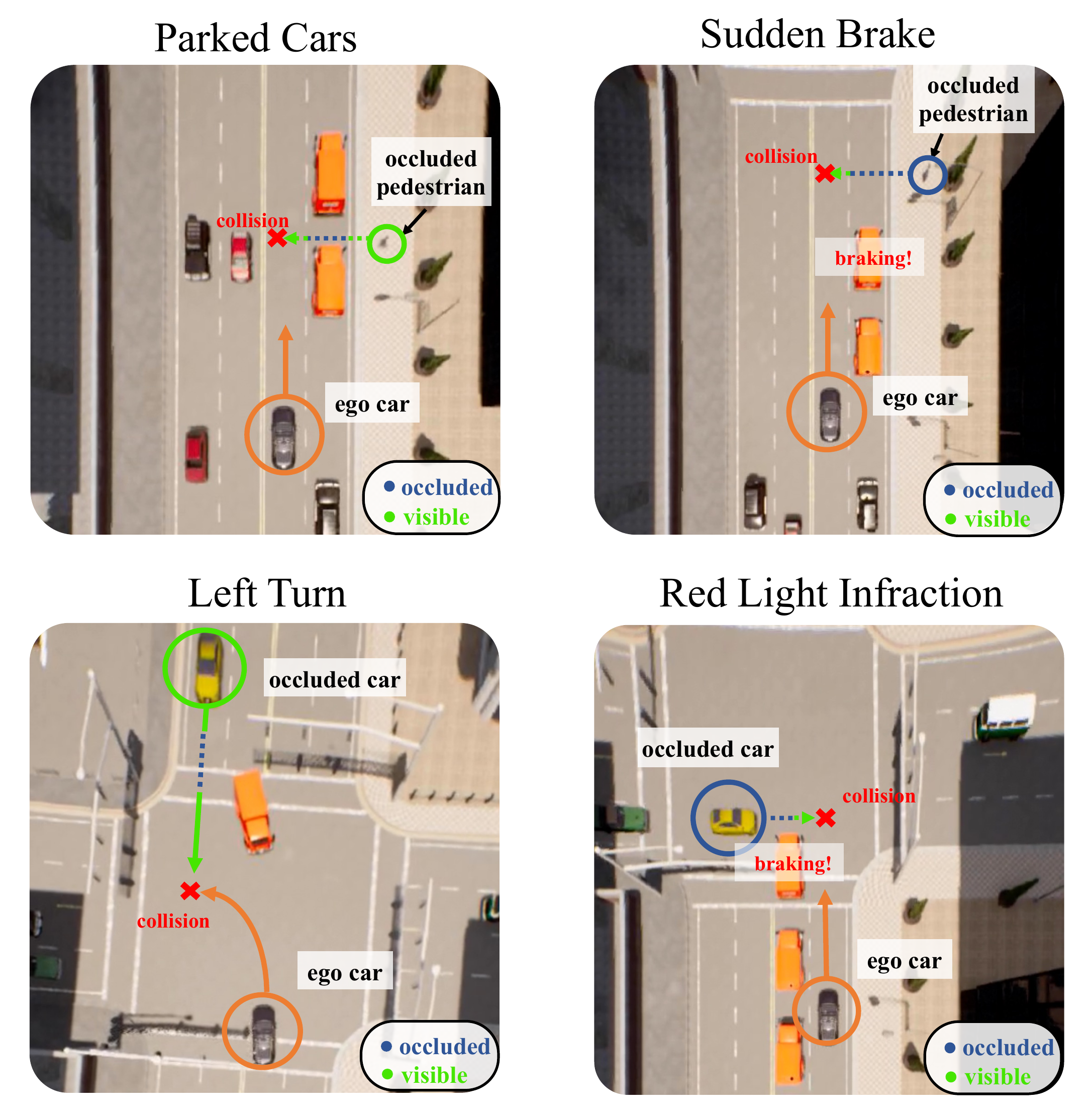}
    \vspace{-1em}
    \caption{An illustration for the four types of occlusion scenarios included in the proposed DOS benchmark. The orange color denotes the ego car. The blue/green dots denote the occluded/visible trajectory of the occluded dangerous object.}
    \label{fig:scenarios}
    \vspace{-1em}
\end{figure}

In order to address the issue that occlusion events are rare in existing datasets and benchmarks, we present the Drive in Occlusion Simulation benchmark (DOS), a CARLA-based framework providing diverse driving scenarios with occluded objects. As shown in Figure~\ref{fig:scenarios}, the proposed DOS benchmark includes four types of challenging occlusion driving scenarios: 

\vspace{0.4em}
\noindent\textbf{Parked Cars} (\#1) The ego vehicle is driving in a straight lane with parked cars on the side. Pedestrians can first appear on the sidewalk (visible) and then suddenly emerge through the occluded areas between parked cars (occluded).

\noindent\textbf{Sudden Brake} (\#2) The ego vehicle is driving in a straight lane along with other vehicles ahead. Pedestrians can suddenly emerge from the sidewalks, causing the other vehicles to brake while remaining invisible to the ego vehicle.

\noindent\textbf{Left Turn} (\#3) The ego vehicle intends to perform an unprotected left turn at an intersection, but a truck in the opposite lane blocks the view of oncoming traffic, intermittently obscuring vehicles driving straight through the intersection.

\noindent\textbf{Red Light Infraction} (\#4) The ego vehicle is crossing an intersection after some trucks. A left-to-right vehicle running a red light suddenly appears, forcing the trucks to brake promptly. But the ego vehicle's view toward the running-light vehicle is blocked by the trucks, so it remains invisible to the ego vehicle.

\vspace{0.4em}

Each of the four scenarios in the DOS benchmark comprises 25 different cases varying in the road environment and background traffic. Compared to a previous occlusion benchmark AUTOCASTSIM~\cite{cui2022coopernaut}, the DOS benchmark: 1) includes occlusions of both vehicles and pedestrians, instead of only vehicles; 2) includes 100 cases of 4 scenarios, instead of only 3 cases of 3 scenarios; 3) considers specific occlusions that can potentially be resolved by temporal reasoning (intermittent occlusion, \#1, \#3) and global reasoning (constant occlusion but with interaction clues, \#2, \#4) about the scene, instead of random occlusions as in AUTOCASTSIM. Thus our scenarios can also serve as a good tracking-with-intermittent-occlusion benchmark and a People-as-Sensor~\cite{afolabi2018people,itkina2022multi} benchmark.

\section{Experiments}
\begin{table}[]
\begin{tabular}{l c c c c}
\toprule
    Rank & Method & DS $\uparrow$ & RC $\uparrow$ & IS $\uparrow$ \\ \hline
    $1$ & ReasonNet (Ours) & $\textbf{79.95}$ & $89.89$ & $\textbf{0.89}$  \\
    $2$ & InterFuser~\cite{shao2022safety}& $76.18$ & $88.23$ & $0.84$  \\
    $3$ & TCP~\cite{wu2022trajectory} & $75.14$ & $85.63$ & $0.87$  \\
    $4$ & LAV~\cite{chen2022learning} & $61.85$ & $\textbf{94.46}$ & $0.64$  \\
    $5$ & TransFuser~\cite{chitta2022transfuser} & $61.18$ & $86.69$ & $0.71$ \\
    $6$ & Latent TransFuser~\cite{chitta2022transfuser} & $45.20$ & $66.31$ & $0.72$ \\
    $7$ & GRIAD~\cite{chekroun2021gri} & $36.79$ & $61.85$ & $0.60$ \\
    $8$ & TransFuser+~\cite{jaeger2021master} & $34.58$ & $69.84$ & $0.56$ \\
    \bottomrule
\end{tabular}
\vspace{-0.5em}
\caption{Performance comparison on the public CARLA leaderboard~\cite{leaderboard} (accessed Nov 2022). For all three metrics, higher is better. Our method ranks first overall on the leaderboard, with the highest driving score (DS) and infraction score (IS), and the second highest route completion (RC). }
\label{table:leaderboard}
\vspace{-1em}
\end{table}

\begin{table*}[]
\centering
\resizebox{0.9\textwidth}{!}{
\begin{tabular}{cccccccccccc}
\toprule
\multicolumn{2}{c}{Setting} & \multicolumn{6}{c}{Town 05 Long} & \multicolumn{4}{c}{DOS} \\ \cmidrule(r){1-2} \cmidrule(r){3-8} \cmidrule(r){9-12}
               $T_{s}$    & $T_{l}$    &  DS $\uparrow$  & RC $\uparrow$   &  IS $\uparrow$   & CR $\downarrow$    & Red $\downarrow$    & Blocked $\downarrow$    &  SR\#1 $\uparrow$   &  SR\#2 $\uparrow$   &  SR\#3 $\uparrow$  &  SR\#4 $\uparrow$   \\ \cmidrule(r){1-2} \cmidrule(r){3-8} \cmidrule(r){9-12}
                0   &   0  &  66.7$\pm$3.8  &\textbf{97.6}$\pm$\textbf{2.7}&  0.68$\pm$0.03  &  0.18$\pm$0.03  & 0.05$\pm$0.02  & \textbf{0.03}$\pm$\textbf{0.03}  & 22$\pm$1.6    &  28$\pm$3.4    &  26$\pm$2.1   &   25$\pm$1.6  \\
                
                1   &   0  &  67.9$\pm$3.4  &96.8$\pm$2.3&  0.70$\pm$0.02  & 0.16$\pm$0.04   &  0.04$\pm$0.03  & 0.05$\pm$0.02  &   30$\pm$3.6   &  38$\pm$3.6    &   32$\pm$2.8  &  32$\pm$3.4   \\ 
                
                2   &   0  &   68.1$\pm$3.1 &96.9$\pm$3.4&  0.70$\pm$0.03  &  0.16$\pm$0.03  &  0.04$\pm$0.02  & 0.05$\pm$0.03   &   28$\pm$5.5   &   48$\pm$4.1   &  38$\pm$4.4   &  52$\pm$3.9   \\
                
                2   &   1  &  70.9$\pm$2.0   &95.7$\pm$3.1&  0.74$\pm$0.02  &  0.13$\pm$0.02  &  0.04$\pm$0.02  & 0.06$\pm$0.04  &  55$\pm$4.4    & 57$\pm$4.1     &   48$\pm$4.1  & 55$\pm$5.5    \\ 
                
                4   &   0  &  70.5$\pm$2.1  &96.4$\pm$2.5& 0.73$\pm$0.04   &  0.14$\pm$0.03  & 0.03$\pm$0.02   & 0.06$\pm$0.03   &   32$\pm$5.4   &   58$\pm$4.4   &   40$\pm$5.5  &  55$\pm$4.9   \\
                
                4   &   2  &  \textbf{73.2}$\pm$\textbf{1.9}  &95.9$\pm$2.3&  \textbf{0.76}$\pm$\textbf{0.03}  &  \textbf{0.11}$\pm$\textbf{0.02}  &  \textbf{0.03}$\pm$\textbf{0.01}  &  0.07$\pm$0.03  &   \textbf{63}$\pm$\textbf{4.2}   &  \textbf{73}$\pm$\textbf{3.6}    &  \textbf{80}$\pm$\textbf{4.2}   &  \textbf{70}$\pm$\textbf{5.5}   \\ 
                        \bottomrule
\end{tabular}
}
    \vspace{-0.5em}
\caption{Ablation study on different short-term buffer size $T_s$ and long-term buffer size $T_l$, on the Town 05 Long benchmark and the proposed DOS benchmark. Performance is evaluated over three runs. CR: Collision rate, Red: Red light violation, Blocked: Vehicle blocked, SR: Success rate. SR\#1 denotes the first kind of scenario in the DOS benchmark. As the two buffer sizes increase, improvement is witnessed in all metrics but the road completion. }
    \vspace{-1em}

\label{table:memory settings}
\end{table*}

\subsection{Experiment Setup}

\noindent\textbf{Implementation}
We implement and evaluate our approach on the open-source CARLA simulator with version 0.9.10.1~\cite{dosovitskiy2017carla}. We use ResNet-50 pretrained on ImageNet as the 2D backbone and PointPillars trained from scratch as the 3D backbone. We predict $T_{f}=4$ time steps for the waypoints and occupancy map, and the interval between each time step is 0.5 seconds.
The memory bank maintains $T_{s}=4$ frames in the short-term buffer and $T_{l}=2$ frames in the long-term buffer. 
The memory bank is updated every $\tau=2$ frame. We refer readers to Appendix \ref{appendix: Implementation Details} for more details.

\noindent\textbf{Dataset Collection}
We collect an expert dataset of 2M frames by running a rule-based expert agent on all 8 public towns and 21 types of weather, with the access to the privileged information in the CARLA simulator. We randomly set routes, spawn dynamic objects and adversarial scenarios provided in~\cite{prakash2021multi}, to diversify the collected data. To ensure the temporal continuity of collected data, the data are collected at a high frequency of 10HZ.

\noindent\textbf{Metrics} We consider three major metrics introduced by the CARLA LeaderBoard: route completion ratio (RC), infraction score (IS), and driving score (DS). The route completion ratio is the percentage of the route completed. The infraction score measures infractions triggered. When collisions or traffic rule violations occur, the infraction score will decay by a discount factor. The driving score is the product of the route completion ratio and the infraction score, describing both driving progress and safety, and thus is the primary ranking metric in the CARLA Leaderboard.

\subsection{Comparison to the state of the art}

Table~\ref{table:leaderboard} shows the top 8 entries on the public CARLA Leaderboard. Readers can refer to Sec~\ref{sec: related work} for descriptions of these methods. Our method outperforms all prior methods, with the highest driving score and infraction score, and the second highest route completion. The previous leading method InterFuser uses a  transformer for sensor fusion but lacks temporal and global reasoning. Compared to InterFuser, our method improved the driving score, road completion, and infraction score by 5\%, 2\%, and 6\% respectively.

\subsection{Ablation study}
\begin{figure}[]
    \centering
    \includegraphics[width=0.45\textwidth]{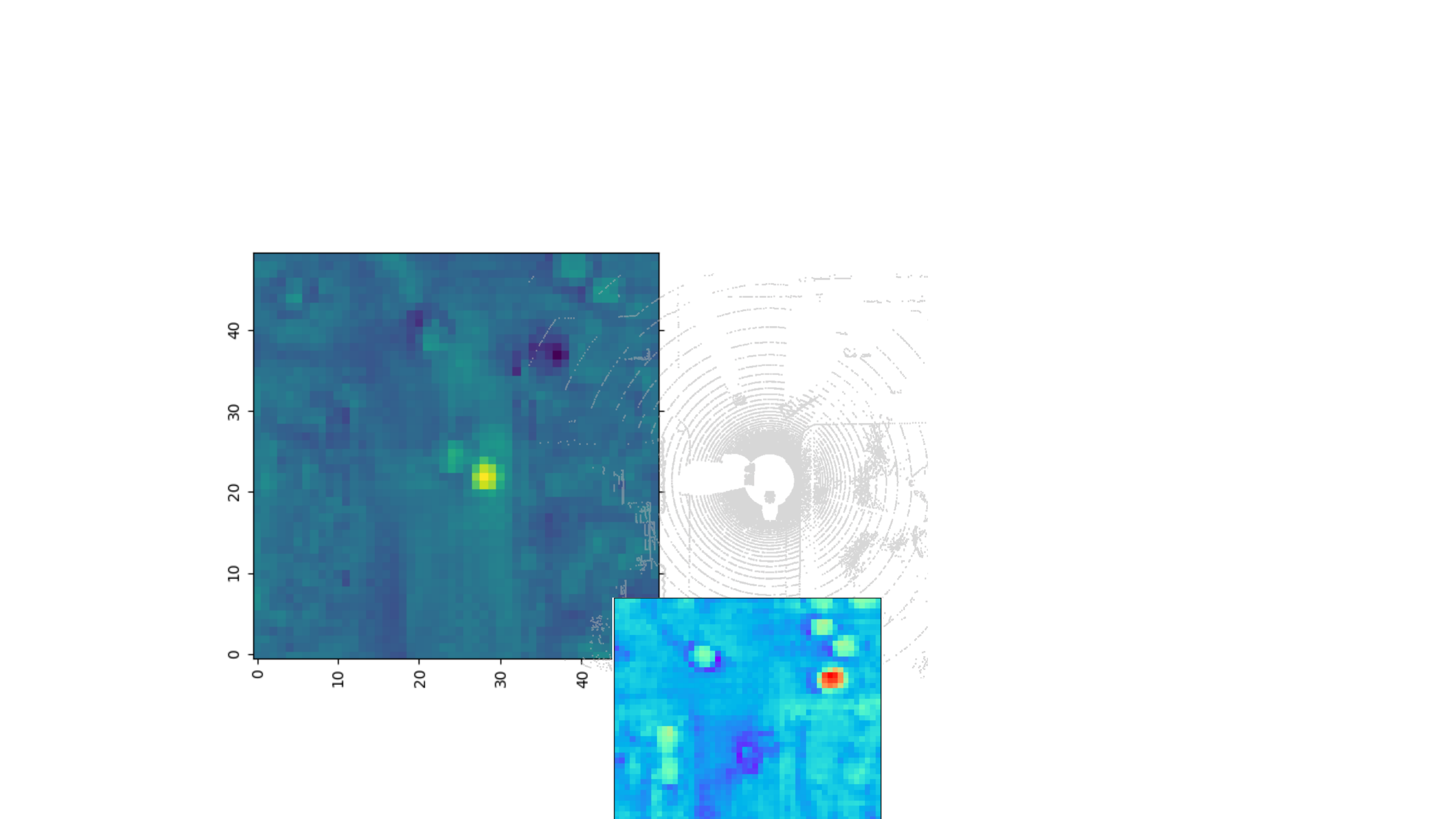}
        \vspace{-0.5em}
    \caption{Visualization of the attention map between one object's current-frame feature query and the historic-frame feature stored in the short-term buffer, in two cases. The object's current-frame feature consistently attends to its corresponding region in the historic feature map.}
    \label{fig:mem attn}
    \vspace{-1em}
\end{figure}

\begin{figure}[]
    \centering
    \includegraphics[width=0.45\textwidth]{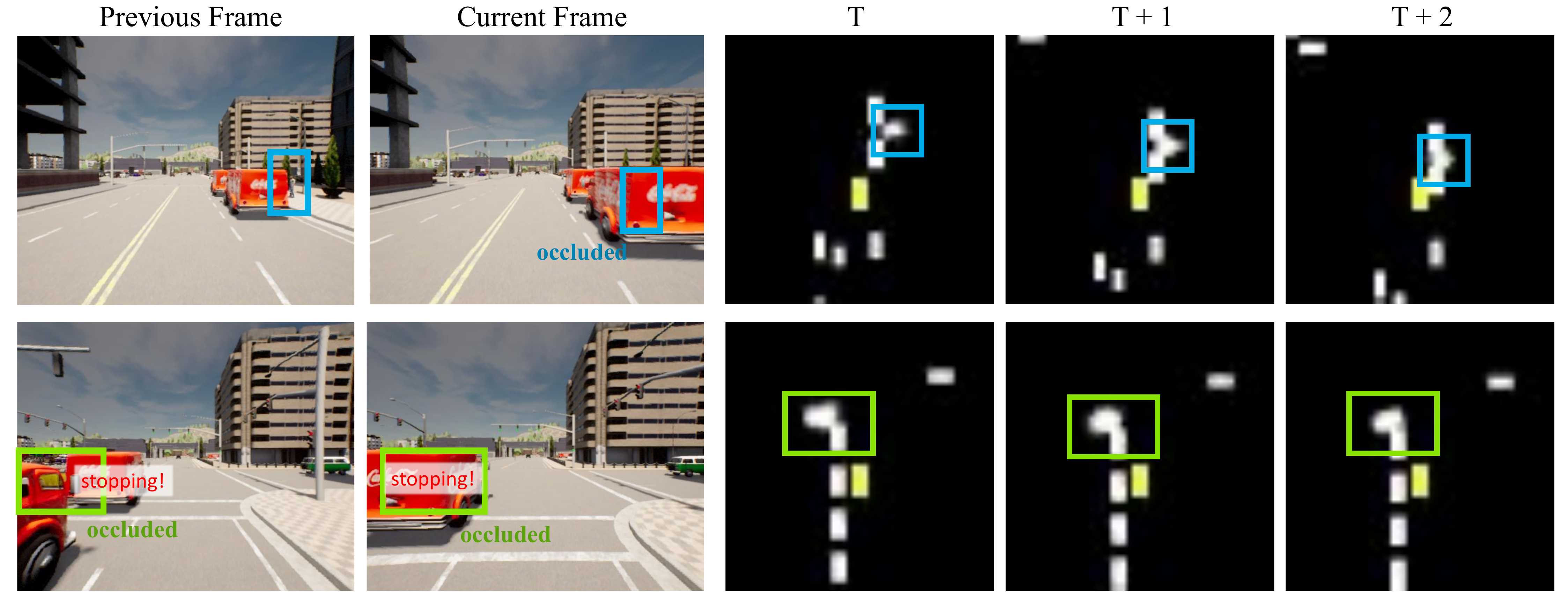}
    \caption{We show two cases of how our framework reasons the presence of the occluded object. 
    In the first case, a pedestrian first appeared on the sidewalk (visible) and then emerges between two parked cars (occluded). 
    In the second case, a vehicle runs the red light, forcing trucks to brake abruptly. But the ego vehicle's view toward the running-light vehicle is blocked by the front trucks, so the running-light vehicle remains invisible to the ego vehicle.
    The rectangles mark the occluded objects.}
        \vspace{-0.5em}
    \label{fig:occlusion}
    \vspace{-1em}
\end{figure}

We investigate the effect of the temporal and global reasoning modules on the Town05 Long benchmark and the DOS benchmark.
For each scenario in DOS, we take 5 cases for training and 20 cases for evaluation. In addition to the three metrics mentioned earlier, we also present four more metrics for detailed analysis: collision rate (CR), red light violation (Red), ego vehicle blocked frequency (Blocked), and success rate (SR). The first three metrics are normalized by the driven distance (km). Visualizations of how the temporal reasoning and global reasoning work can be found at Figure~\ref{fig:mem attn} and Figure~\ref{fig:occlusion} respectively.

\noindent\textbf{Memory Size} Table~\ref{table:memory settings} studies the effect of different short-term buffer size $T_s$ and long-term buffer size $T_l$. The overall observation is that, as the two buffer sizes increase, improvement is witnessed in all metrics but road completion. Specifically, when the long-term memory is removed ($T_{l}=0$), the average success rates drop sharply from 71.5 to 36 on DOS scenarios that require keeping track of intermittently occluded objects (\#1 and \#3). If we remove the temporal reasoning module ($T_{s} = T_{l}=0$), the driving score on the Town05 benchmark drops by 9\%, and the average success rate on the DOS benchmark drops by 46\%. We hypothesize that the drop in performance is because 1) it can be really hard to accurately estimate the objects' future motion based only on single-frame data; 2) temporal information can help keep track of objects that are intermittently occluded; 3) the global reasoning module may also work poorly when historic information is missing.

\begin{table*}[t]
\centering
\resizebox{0.9\textwidth}{!}{
\begin{tabular}{lcccccccccc}
\toprule
\multirow{2}{*}{Setting} & \multicolumn{6}{c}{Town 05 Long} & \multicolumn{4}{c}{DOS} \\ \cmidrule(r){2-7} \cmidrule(r){8-11}
                        &  DS $\uparrow$  & RC $\uparrow$   &  IS $\uparrow$   & CR $\downarrow$    & Red $\downarrow$    & Blocked $\downarrow$    &  SR\#1 $\uparrow$   &  SR\#2 $\uparrow$   &  SR\#3 $\uparrow$  &  SR\#4 $\uparrow$   \\ \cmidrule(r){1-1} \cmidrule(r){2-7} \cmidrule(r){8-11}
                  Random      &  71.2$\pm$5.4  &  \textbf{96.6}$\pm$\textbf{2.4}  & 0.74$\pm$0.04   & 0.13$\pm$0.04   &  0.03$\pm$0.01  &  0.06$\pm$0.02  &  33$\pm$4.4    &  55$\pm$6.2    &  42$\pm$5.5   &  53$\pm$5.4  \\
                  Usage-based      &  72.0$\pm$3.9  & 95.9$\pm$2.2   & 0.75$\pm$0.04   & 0.12$\pm$0.03   &  0.03$\pm$0.01  & 0.06$\pm$0.02   & 45$\pm$4.2     &  62$\pm$3.4    &  53$\pm$2.8  & 62$\pm$3.9    \\ 
                  Object-based      &  72.2$\pm$3.7  &  96.1$\pm$3.0  & 0.75$\pm$0.03   & 0.12$\pm$0.03   &  0.03$\pm$0.01  & \textbf{0.05}$\pm$\textbf{0.02}   &  57$\pm$4.1    &  65$\pm$3.6    &  73$\pm$4.4   &  60$\pm$3.7   \\ 
                  Full (Ours)     &  \textbf{73.2}$\pm$\textbf{1.9}  &95.9$\pm$2.3&  \textbf{0.76}$\pm$\textbf{0.03}  &  \textbf{0.11}$\pm$\textbf{0.02}  &  \textbf{0.03}$\pm$\textbf{0.01}  &  0.07$\pm$0.03  &   \textbf{63}$\pm$\textbf{4.2}   &  \textbf{73}$\pm$\textbf{3.6}    &  \textbf{80}$\pm$\textbf{4.2}   &  \textbf{70}$\pm$\textbf{5.5}    \\ 
                        \bottomrule
\end{tabular}}
\vspace{-0.5em}
\caption{Ablation study on different long-term memory selection strategies. Our proposed strategy considering both the usage and object criteria outperforms the random selection strategy and the two methods with only one criteria, especially on the DOS benchmark.}
\vspace{-0.5em}
\label{table:memory selection}
\end{table*}

\begin{table*}[t]
\centering
\resizebox{0.9\textwidth}{!}{
\begin{tabular}{lcccccccccc}
\toprule
\multirow{2}{*}{Setting} & \multicolumn{6}{c}{Town 05 Long} & \multicolumn{4}{c}{DOS} \\ \cmidrule(r){2-7} \cmidrule(r){8-11}
                        &  DS $\uparrow$  & RC $\uparrow$   &  IS $\uparrow$   & CR $\downarrow$    & Red $\downarrow$    & Blocked $\downarrow$    &  SR\#1 $\uparrow$   &  SR\#2 $\uparrow$   &  SR\#3 $\uparrow$  &  SR\#4 $\uparrow$   \\ \cmidrule(r){1-1} \cmidrule(r){2-7} \cmidrule(r){8-11}
            No global reasoning      & 68.9$\pm$4.6   & \textbf{97.4}$\pm$\textbf{2.9}   &  0.71$\pm$0.04  &  0.15$\pm$0.04  &   0.05$\pm$0.02  &  \textbf{0.05}$\pm$\textbf{0.02}  &  28$\pm$2.8    &   34$\pm$3.4   &  29$\pm$2.0   &  27$\pm$3.6   \\  
            
           No consistency loss  & 72.2$\pm$3.4   & 96.1$\pm$3.2   & 0.75$\pm$0.03   &  0.12$\pm$0.02  &  \textbf{0.03}$\pm$\textbf{0.02}  & 0.06$\pm$0.03   &  60$\pm$4.1    &   72$\pm$3.9   &  77$\pm$4.9   &  68$\pm$4.2   \\
           No traffic sign prediction   & 71.1$\pm$2.7   & 96.0$\pm$4.1   & 0.74$\pm$0.03   &  \textbf{0.11}$\pm$\textbf{0.03}  &  0.05$\pm$0.03  & 0.07$\pm$0.03   &  62$\pm$4.4    & 72$\pm$4.0     &  \textbf{82}$\pm$\textbf{2.8}   &  \textbf{70}$\pm$\textbf{4.1}   \\ 
                    
           Full (Ours)     &  \textbf{73.2}$\pm$\textbf{1.9}  &95.9$\pm$2.3&  \textbf{0.76}$\pm$\textbf{0.03}  &  \textbf{0.11}$\pm$\textbf{0.02}  &  \textbf{0.03}$\pm$\textbf{0.01}  &  0.07$\pm$0.03  &   \textbf{63}$\pm$\textbf{4.2}   &  \textbf{73}$\pm$\textbf{3.6}    &  80$\pm$4.2   &  \textbf{70}$\pm$\textbf{5.5}    \\ 
           \bottomrule
\end{tabular}}
\vspace{-0.5em}
\caption{Ablation study on the global reasoning module. The performance would drop when 1) the entire global reasoning module is removed; 2) the consistency loss is not applied; 3) the traffic sign feature from the reasoning module is not utilized.}
\vspace{-1em}
\label{table:global reasoning}
\end{table*}

\noindent\textbf{Long-Term Memory Selection Strategy} Table~\ref{table:memory selection} studies the performance of different long-term memory selection strategies. Specifically, the proposed strategy in Sec~\ref{sec:temporal reasoning} includes two selection criteria. So here we ablate the effect of the two criteria by 1) only selecting the short-term feature with top-\textit{K} usages (usage-based); 2) only selecting the feature with a high probability of the existence of an object (object-based). Besides, we also compared a random selection strategy. As in Table~\ref{table:memory selection}, the random selection strategy has the poorest performance especially on the DOS benchmark, as random selection could miss important and representative features on the scene. Compared to our strategy utilizing both criteria, the two ablations omitting one of the criteria have a performance drop, especially on the DOS benchmark. The usage-based strategy performs worse than the object-based, showing that the features of objects could be more informative for capturing historic behaviors.

\noindent\textbf{Global Reasoning Design} Table~\ref{table:global reasoning} studies the performance when different designs of the global reasoning module are applied. First, we remove the entire module and observe a significant drop in all metrics but the road completion. For instance, the average success rate on the DOS benchmark dropped from 71.5 to 29.5. This demonstrates the effectiveness of global reasoning, especially in occlusion events. Second, we ablated the consistency loss, which could alleviate the sub-optimal issues in collected expert data. A removal of consistency loss leads to a lower driving score and higher collision rate on the Town 05 benchmark and a lower success rate on the DOS benchmark.
Third, excluding the traffic sign feature from the global reasoning model results in an increase on the red light violation. One explanation is that the traffic sign feature from the global reasoning module could help reason the distant traffic light state according to other road participants' behavior.

\section{Conclusion}
We present ReasonNet, a novel end-to-end autonomous driving framework including two major components: a temporal reasoning module and a global reasoning module. The temporal reasoning module processes the historic information on the driving scene for high-fidelity forecasting of other road participants and dynamically maintains a temporal memory bank. The global reasoning module models the interaction and relationship among the objects and environment to detect adverse events, especially occlusion, and improve overall perception performance. Our method pushes the state-of-the-art performance of the CARLA leaderboard by a considerable margin. Moreover, we also publicly release a new benchmark DOS consisting of diverse occlusion scenarios, to facilitate the study of occlusion detection in the field of end-to-end autonomous driving.

{\small
\bibliographystyle{ieee_fullname}
\bibliography{ref}

\begin{thebibliography}{10}\itemsep=-1pt

\bibitem{jaeger2021master}
Expert drivers for autonomous driving.
\newblock
  url{https://kait0.github.io/files/master\_thesis\_bernhard\_jaeger.pdf},
  2021.

\bibitem{afolabi2018people}
Oladapo Afolabi, Katherine Driggs-Campbell, Roy Dong, Mykel~J Kochenderfer, and
  S~Shankar Sastry.
\newblock People as sensors: Imputing maps from human actions.
\newblock In {\em 2018 IEEE/RSJ International Conference on Intelligent Robots
  and Systems (IROS)}, pages 2342--2348. IEEE, 2018.

\bibitem{ba2016layer}
Jimmy~Lei Ba, Jamie~Ryan Kiros, and Geoffrey~E Hinton.
\newblock Layer normalization.
\newblock {\em arXiv preprint arXiv:1607.06450}, 2016.

\bibitem{tesla}
Neal~E. Boudette.
\newblock Teslas self-driving system cleared in deadly crash.
\newblock The New York Times, 2017.

\bibitem{cao2022observation}
Jinkun Cao, Xinshuo Weng, Rawal Khirodkar, Jiangmiao Pang, and Kris Kitani.
\newblock Observation-centric sort: Rethinking sort for robust multi-object
  tracking.
\newblock {\em arXiv preprint arXiv:2203.14360}, 2022.

\bibitem{casas2020implicit}
Sergio Casas, Cole Gulino, Simon Suo, Katie Luo, Renjie Liao, and Raquel
  Urtasun.
\newblock Implicit latent variable model for scene-consistent motion
  forecasting.
\newblock In {\em Computer Vision--ECCV 2020: 16th European Conference,
  Glasgow, UK, August 23--28, 2020, Proceedings, Part XXIII 16}, pages
  624--641. Springer, 2020.

\bibitem{casas2018intentnet}
Sergio Casas, Wenjie Luo, and Raquel Urtasun.
\newblock Intentnet: Learning to predict intention from raw sensor data.
\newblock In {\em Conference on Robot Learning}, pages 947--956. PMLR, 2018.

\bibitem{chekroun2021gri}
Raphael Chekroun, Marin Toromanoff, Sascha Hornauer, and Fabien Moutarde.
\newblock Gri: General reinforced imitation and its application to vision-based
  autonomous driving.
\newblock {\em arXiv preprint arXiv:2111.08575}, 2021.

\bibitem{chen2021learning}
Dian Chen, Vladlen Koltun, and Philipp Kr{\"a}henb{\"u}hl.
\newblock Learning to drive from a world on rails.
\newblock In {\em Proceedings of the IEEE/CVF International Conference on
  Computer Vision}, pages 15590--15599, 2021.

\bibitem{chen2022learning}
Dian Chen and Philipp Kr{\"a}henb{\"u}hl.
\newblock Learning from all vehicles.
\newblock In {\em Proceedings of the IEEE/CVF Conference on Computer Vision and
  Pattern Recognition}, pages 17222--17231, 2022.

\bibitem{chen2020learning}
Dian Chen, Brady Zhou, Vladlen Koltun, and Philipp Kr{\"a}henb{\"u}hl.
\newblock Learning by cheating.
\newblock In {\em Conference on Robot Learning}, pages 66--75. PMLR, 2020.

\bibitem{cheng2022xmem}
Ho~Kei Cheng and Alexander~G Schwing.
\newblock Xmem: Long-term video object segmentation with an atkinson-shiffrin
  memory model.
\newblock In {\em European Conference on Computer Vision}, pages 640--658.
  Springer, 2022.

\bibitem{cheng2021rethinking}
Ho~Kei Cheng, Yu-Wing Tai, and Chi-Keung Tang.
\newblock Rethinking space-time networks with improved memory coverage for
  efficient video object segmentation.
\newblock {\em Advances in Neural Information Processing Systems},
  34:11781--11794, 2021.

\bibitem{chitta2021neat}
Kashyap Chitta, Aditya Prakash, and Andreas Geiger.
\newblock Neat: Neural attention fields for end-to-end autonomous driving.
\newblock In {\em Proceedings of the IEEE/CVF International Conference on
  Computer Vision}, pages 15793--15803, 2021.

\bibitem{chitta2022transfuser}
Kashyap Chitta, Aditya Prakash, Bernhard Jaeger, Zehao Yu, Katrin Renz, and
  Andreas Geiger.
\newblock Transfuser: Imitation with transformer-based sensor fusion for
  autonomous driving.
\newblock {\em arXiv preprint arXiv:2205.15997}, 2022.

\bibitem{cho2014learning}
Kyunghyun Cho, Bart Van~Merri{\"e}nboer, Caglar Gulcehre, Dzmitry Bahdanau,
  Fethi Bougares, Holger Schwenk, and Yoshua Bengio.
\newblock Learning phrase representations using rnn encoder-decoder for
  statistical machine translation.
\newblock {\em arXiv preprint arXiv:1406.1078}, 2014.

\bibitem{codevilla2018end}
Felipe Codevilla, Matthias M{\"u}ller, Antonio L{\'o}pez, Vladlen Koltun, and
  Alexey Dosovitskiy.
\newblock End-to-end driving via conditional imitation learning.
\newblock In {\em 2018 IEEE International Conference on Robotics and Automation
  (ICRA)}, pages 4693--4700. IEEE, 2018.

\bibitem{codevilla2019exploring}
Felipe Codevilla, Eder Santana, Antonio~M L{\'o}pez, and Adrien Gaidon.
\newblock Exploring the limitations of behavior cloning for autonomous driving.
\newblock In {\em Proceedings of the IEEE/CVF International Conference on
  Computer Vision}, pages 9329--9338, 2019.

\bibitem{cui2022coopernaut}
Jiaxun Cui, Hang Qiu, Dian Chen, Peter Stone, and Yuke Zhu.
\newblock Coopernaut: End-to-end driving with cooperative perception for
  networked vehicles.
\newblock In {\em Proceedings of the IEEE/CVF Conference on Computer Vision and
  Pattern Recognition}, pages 17252--17262, 2022.

\bibitem{dosovitskiy2020image}
Alexey Dosovitskiy, Lucas Beyer, Alexander Kolesnikov, Dirk Weissenborn,
  Xiaohua Zhai, Thomas Unterthiner, Mostafa Dehghani, Matthias Minderer, Georg
  Heigold, Sylvain Gelly, et~al.
\newblock An image is worth 16x16 words: Transformers for image recognition at
  scale.
\newblock {\em arXiv preprint arXiv:2010.11929}, 2020.

\bibitem{dosovitskiy2017carla}
Alexey Dosovitskiy, German Ros, Felipe Codevilla, Antonio Lopez, and Vladlen
  Koltun.
\newblock Carla: An open urban driving simulator.
\newblock In {\em Conference on robot learning}, pages 1--16. PMLR, 2017.

\bibitem{feichtenhofer2019slowfast}
Christoph Feichtenhofer, Haoqi Fan, Jitendra Malik, and Kaiming He.
\newblock Slowfast networks for video recognition.
\newblock In {\em Proceedings of the IEEE/CVF international conference on
  computer vision}, pages 6202--6211, 2019.

\bibitem{gao2020vectornet}
Jiyang Gao, Chen Sun, Hang Zhao, Yi Shen, Dragomir Anguelov, Congcong Li, and
  Cordelia Schmid.
\newblock Vectornet: Encoding hd maps and agent dynamics from vectorized
  representation.
\newblock In {\em Proceedings of the IEEE/CVF Conference on Computer Vision and
  Pattern Recognition}, pages 11525--11533, 2020.

\bibitem{uber}
Samuel Gibbs.
\newblock Ubers self-driving car saw the pedestrian but didnt swerve–report.
\newblock The Guardian, 2018.

\bibitem{gou2022driver}
Chao Gou, Yuchen Zhou, and Dan Li.
\newblock Driver attention prediction based on convolution and transformers.
\newblock {\em The Journal of Supercomputing}, 78(6):8268--8284, 2022.

\bibitem{gu2021densetnt}
Junru Gu, Chen Sun, and Hang Zhao.
\newblock Densetnt: End-to-end trajectory prediction from dense goal sets.
\newblock In {\em Proceedings of the IEEE/CVF International Conference on
  Computer Vision}, pages 15303--15312, 2021.

\bibitem{he2016deep}
Kaiming He, Xiangyu Zhang, Shaoqing Ren, and Jian Sun.
\newblock Deep residual learning for image recognition.
\newblock In {\em Proceedings of the IEEE conference on computer vision and
  pattern recognition}, pages 770--778, 2016.

\bibitem{hu2021fiery}
Anthony Hu, Zak Murez, Nikhil Mohan, Sof{\'\i}a Dudas, Jeffrey Hawke, Vijay
  Badrinarayanan, Roberto Cipolla, and Alex Kendall.
\newblock Fiery: future instance prediction in bird's-eye view from surround
  monocular cameras.
\newblock In {\em Proceedings of the IEEE/CVF International Conference on
  Computer Vision}, pages 15273--15282, 2021.

\bibitem{hu2022st}
Shengchao Hu, Li Chen, Penghao Wu, Hongyang Li, Junchi Yan, and Dacheng Tao.
\newblock St-p3: End-to-end vision-based autonomous driving via
  spatial-temporal feature learning.
\newblock In {\em Computer Vision--ECCV 2022: 17th European Conference, Tel
  Aviv, Israel, October 23--27, 2022, Proceedings, Part XXXVIII}, pages
  533--549. Springer, 2022.

\bibitem{ioffe2015batch}
Sergey Ioffe and Christian Szegedy.
\newblock Batch normalization: Accelerating deep network training by reducing
  internal covariate shift.
\newblock In {\em International conference on machine learning}, pages
  448--456. PMLR, 2015.

\bibitem{itkina2022multi}
Masha Itkina, Ye-Ji Mun, Katherine Driggs-Campbell, and Mykel~J Kochenderfer.
\newblock Multi-agent variational occlusion inference using people as sensors.
\newblock In {\em 2022 International Conference on Robotics and Automation
  (ICRA)}, pages 4585--4591. IEEE, 2022.

\bibitem{jia2023towards}
Xiaosong Jia, Li Chen, Penghao Wu, Jia Zeng, Junchi Yan, Hongyang Li, and Yu
  Qiao.
\newblock Towards capturing the temporal dynamics for trajectory prediction: a
  coarse-to-fine approach.
\newblock In {\em Conference on Robot Learning}, pages 910--920. PMLR, 2023.

\bibitem{jia2022hdgt}
Xiaosong Jia, Penghao Wu, Li Chen, Hongyang Li, Yu Liu, and Junchi Yan.
\newblock Hdgt: Heterogeneous driving graph transformer for multi-agent
  trajectory prediction via scene encoding.
\newblock {\em arXiv preprint arXiv:2205.09753}, 2022.

\bibitem{kim2019grounding}
Jinkyu Kim, Teruhisa Misu, Yi-Ting Chen, Ashish Tawari, and John Canny.
\newblock Grounding human-to-vehicle advice for self-driving vehicles.
\newblock In {\em Proceedings of the IEEE/CVF Conference on Computer Vision and
  Pattern Recognition}, pages 10591--10599, 2019.

\bibitem{lang2019pointpillars}
Alex~H Lang, Sourabh Vora, Holger Caesar, Lubing Zhou, Jiong Yang, and Oscar
  Beijbom.
\newblock Pointpillars: Fast encoders for object detection from point clouds.
\newblock In {\em Proceedings of the IEEE/CVF conference on computer vision and
  pattern recognition}, pages 12697--12705, 2019.

\bibitem{li2020end}
Lingyun~Luke Li, Bin Yang, Ming Liang, Wenyuan Zeng, Mengye Ren, Sean Segal,
  and Raquel Urtasun.
\newblock End-to-end contextual perception and prediction with interaction
  transformer.
\newblock In {\em 2020 IEEE/RSJ International Conference on Intelligent Robots
  and Systems (IROS)}, pages 5784--5791. IEEE, 2020.

\bibitem{li2022bevformer}
Zhiqi Li, Wenhai Wang, Hongyang Li, Enze Xie, Chonghao Sima, Tong Lu, Qiao Yu,
  and Jifeng Dai.
\newblock Bevformer: Learning bird's-eye-view representation from multi-camera
  images via spatiotemporal transformers.
\newblock {\em arXiv preprint arXiv:2203.17270}, 2022.

\bibitem{lian2022monojsg}
Qing Lian, Peiliang Li, and Xiaozhi Chen.
\newblock Monojsg: Joint semantic and geometric cost volume for monocular 3d
  object detection.
\newblock In {\em Proceedings of the IEEE/CVF Conference on Computer Vision and
  Pattern Recognition}, pages 1070--1079, 2022.

\bibitem{lian2022semi}
Qing Lian, Yanbo Xu, Weilong Yao, Yingcong Chen, and Tong Zhang.
\newblock Semi-supervised monocular 3d object detection by multi-view
  consistency.
\newblock In {\em Computer Vision--ECCV 2022: 17th European Conference, Tel
  Aviv, Israel, October 23--27, 2022, Proceedings, Part VIII}, pages 715--731.
  Springer, 2022.

\bibitem{lian2022exploring}
Qing Lian, Botao Ye, Ruijia Xu, Weilong Yao, and Tong Zhang.
\newblock Exploring geometric consistency for monocular 3d object detection.
\newblock In {\em Proceedings of the IEEE/CVF Conference on Computer Vision and
  Pattern Recognition}, pages 1685--1694, 2022.

\bibitem{liang2020pnpnet}
Ming Liang, Bin Yang, Wenyuan Zeng, Yun Chen, Rui Hu, Sergio Casas, and Raquel
  Urtasun.
\newblock Pnpnet: End-to-end perception and prediction with tracking in the
  loop.
\newblock In {\em Proceedings of the IEEE/CVF Conference on Computer Vision and
  Pattern Recognition}, pages 11553--11562, 2020.

\bibitem{liu2022bevfusion}
Zhijian Liu, Haotian Tang, Alexander Amini, Xinyu Yang, Huizi Mao, Daniela Rus,
  and Song Han.
\newblock Bevfusion: Multi-task multi-sensor fusion with unified bird's-eye
  view representation.
\newblock {\em arXiv preprint arXiv:2205.13542}, 2022.

\bibitem{loshchilov2016sgdr}
Ilya Loshchilov and Frank Hutter.
\newblock Sgdr: Stochastic gradient descent with warm restarts.
\newblock {\em arXiv preprint arXiv:1608.03983}, 2016.

\bibitem{loshchilov2018decoupled}
Ilya Loshchilov and Frank Hutter.
\newblock Decoupled weight decay regularization.
\newblock In {\em International Conference on Learning Representations}, 2018.

\bibitem{meinhardt2022trackformer}
Tim Meinhardt, Alexander Kirillov, Laura Leal-Taixe, and Christoph
  Feichtenhofer.
\newblock Trackformer: Multi-object tracking with transformers.
\newblock In {\em Proceedings of the IEEE/CVF Conference on Computer Vision and
  Pattern Recognition}, pages 8844--8854, 2022.

\bibitem{patrick2021keeping}
Mandela Patrick, Dylan Campbell, Yuki Asano, Ishan Misra, Florian Metze,
  Christoph Feichtenhofer, Andrea Vedaldi, and Jo{\~a}o~F Henriques.
\newblock Keeping your eye on the ball: Trajectory attention in video
  transformers.
\newblock {\em Advances in neural information processing systems},
  34:12493--12506, 2021.

\bibitem{prakash2021multi}
Aditya Prakash, Kashyap Chitta, and Andreas Geiger.
\newblock Multi-modal fusion transformer for end-to-end autonomous driving.
\newblock In {\em Proceedings of the IEEE/CVF Conference on Computer Vision and
  Pattern Recognition}, pages 7077--7087, 2021.

\bibitem{qi2018frustum}
Charles~R Qi, Wei Liu, Chenxia Wu, Hao Su, and Leonidas~J Guibas.
\newblock Frustum pointnets for 3d object detection from rgb-d data.
\newblock In {\em Proceedings of the IEEE conference on computer vision and
  pattern recognition}, pages 918--927, 2018.

\bibitem{qian2021blending}
Shengju Qian, Hao Shao, Yi Zhu, Mu Li, and Jiaya Jia.
\newblock Blending anti-aliasing into vision transformer.
\newblock {\em Advances in Neural Information Processing Systems},
  34:5416--5429, 2021.

\bibitem{shao2020temporal}
Hao Shao, Shengju Qian, and Yu Liu.
\newblock Temporal interlacing network.
\newblock In {\em Proceedings of the AAAI Conference on Artificial
  Intelligence}, volume~34, pages 11966--11973, 2020.

\bibitem{shao2022safety}
Hao Shao, Letian Wang, Ruobing Chen, Hongsheng Li, and Yu Liu.
\newblock Safety-enhanced autonomous driving using interpretable sensor fusion
  transformer.
\newblock {\em arXiv preprint arXiv:2207.14024}, 2022.

\bibitem{sun2020transtrack}
Peize Sun, Jinkun Cao, Yi Jiang, Rufeng Zhang, Enze Xie, Zehuan Yuan, Changhu
  Wang, and Ping Luo.
\newblock Transtrack: Multiple object tracking with transformer.
\newblock {\em arXiv preprint arXiv:2012.15460}, 2020.

\bibitem{leaderboard}
CARLA team.
\newblock Carla autonomous driving leaderboard.
\newblock \url{https://leaderboard.carla.org/}, 2020.
\newblock Accessed: 2021-02-11.

\bibitem{toromanoff2020end}
Marin Toromanoff, Emilie Wirbel, and Fabien Moutarde.
\newblock End-to-end model-free reinforcement learning for urban driving using
  implicit affordances.
\newblock In {\em Proceedings of the IEEE/CVF conference on computer vision and
  pattern recognition}, pages 7153--7162, 2020.

\bibitem{vaswani2017attention}
Ashish Vaswani, Noam Shazeer, Niki Parmar, Jakob Uszkoreit, Llion Jones,
  Aidan~N Gomez, {\L}ukasz Kaiser, and Illia Polosukhin.
\newblock Attention is all you need.
\newblock {\em Advances in neural information processing systems}, 30, 2017.

\bibitem{velickovic2017graph}
Petar Velickovic, Guillem Cucurull, Arantxa Casanova, Adriana Romero, Pietro
  Lio, and Yoshua Bengio.
\newblock Graph attention networks.
\newblock {\em stat}, 1050:20, 2017.

\bibitem{vora2020pointpainting}
Sourabh Vora, Alex~H Lang, Bassam Helou, and Oscar Beijbom.
\newblock Pointpainting: Sequential fusion for 3d object detection.
\newblock In {\em Proceedings of the IEEE/CVF conference on computer vision and
  pattern recognition}, pages 4604--4612, 2020.

\bibitem{wang2021hierarchical}
Letian Wang, Yeping Hu, Liting Sun, Wei Zhan, Masayoshi Tomizuka, and Changliu
  Liu.
\newblock Hierarchical adaptable and transferable networks (hatn) for driving
  behavior prediction.
\newblock {\em arXiv preprint arXiv:2111.00788}, 2021.

\bibitem{wang2022transferable}
Letian Wang, Yeping Hu, Liting Sun, Wei Zhan, Masayoshi Tomizuka, and Changliu
  Liu.
\newblock Transferable and adaptable driving behavior prediction.
\newblock {\em arXiv preprint arXiv:2202.05140}, 2022.

\bibitem{wang2021socially}
Letian Wang, Liting Sun, Masayoshi Tomizuka, and Wei Zhan.
\newblock Socially-compatible behavior design of autonomous vehicles with
  verification on real human data.
\newblock {\em IEEE Robotics and Automation Letters}, 6(2):3421--3428, 2021.

\bibitem{wei2021perceive}
Bob Wei, Mengye Ren, Wenyuan Zeng, Ming Liang, Bin Yang, and Raquel Urtasun.
\newblock Perceive, attend, and drive: Learning spatial attention for safe
  self-driving.
\newblock In {\em 2021 IEEE International Conference on Robotics and Automation
  (ICRA)}, pages 4875--4881. IEEE, 2021.

\bibitem{wu2020motionnet}
Pengxiang Wu, Siheng Chen, and Dimitris~N Metaxas.
\newblock Motionnet: Joint perception and motion prediction for autonomous
  driving based on bird's eye view maps.
\newblock In {\em Proceedings of the IEEE/CVF conference on computer vision and
  pattern recognition}, pages 11385--11395, 2020.

\bibitem{wu2022trajectory}
Penghao Wu, Xiaosong Jia, Li Chen, Junchi Yan, Hongyang Li, and Yu Qiao.
\newblock Trajectory-guided control prediction for end-to-end autonomous
  driving: A simple yet strong baseline.
\newblock {\em arXiv preprint arXiv:2206.08129}, 2022.

\bibitem{yin2021center}
Tianwei Yin, Xingyi Zhou, and Philipp Krahenbuhl.
\newblock Center-based 3d object detection and tracking.
\newblock In {\em Proceedings of the IEEE/CVF conference on computer vision and
  pattern recognition}, pages 11784--11793, 2021.

\bibitem{zhang2022bytetrack}
Yifu Zhang, Peize Sun, Yi Jiang, Dongdong Yu, Fucheng Weng, Zehuan Yuan, Ping
  Luo, Wenyu Liu, and Xinggang Wang.
\newblock Bytetrack: Multi-object tracking by associating every detection box.
\newblock In {\em European Conference on Computer Vision}, pages 1--21.
  Springer, 2022.

\bibitem{zhang2021end}
Zhejun Zhang, Alexander Liniger, Dengxin Dai, Fisher Yu, and Luc Van~Gool.
\newblock End-to-end urban driving by imitating a reinforcement learning coach.
\newblock In {\em Proceedings of the IEEE/CVF International Conference on
  Computer Vision}, pages 15222--15232, 2021.

\end{thebibliography}
}

\newpage
\appendix

\clearpage

\begin{table*}[t!]
\centering
\scalebox{0.85}{
\begin{tabular}{l@{\ }c@{\ \ \ }c@{\ }c@{\ }c@{\ }c@{\ }c@{\ }c@{\ }c@{\ }c@{\ }c@{\ }}
    \toprule
     \thead{Rank} & \thead{Method} & \thead{Driving \\ Score } & \thead{Route \\ Completion} & \thead{Infraction \\ Score} & \thead{Vehicle \\ Collisions} & \thead{Pedestrian \\ Collisions} & \thead{Layout \\ Collisions} & \thead{Red light \\ Violations} & \thead{Offroad \\ Infractions} & \thead{Blocked \\ Infractions} \\
    \cmidrule(r){1-2}
    \cmidrule(r){3-11}
    $1$ & ReasonNet (Ours)& \textbf{79.95} & 89.89 & \textbf{0.89} & \textbf{0.13} & 0.02  & 0.01 & 0.08 & \textbf{0.04} & 0.33\\
    $2$ & InterFuser~\cite{shao2022safety} & 76.18 & 88.23 & 0.84 & 0.37 & 0.04  & 0.14 & 0.22 & 0.13 & 0.43	\\
    $3$ & TCP~\cite{wu2022trajectory} & 75.14 & 85.63 & 0.87 & 0.32 & \textbf{0.00}  & \textbf{0.00} & 0.09 & \textbf{0.04} & 0.54 \\
    $4$ & LAV~\cite{chen2022learning} & 61.85 & \textbf{94.46} & 0.64 & 0.70 & 0.04 & 0.02 & 0.17 & 0.25 & \textbf{0.10} \\
    $5$ & TransFuser~\cite{chitta2022transfuser} & 61.18	 & 86.69 & 0.04 & 0.71 & 0.81 & 0.01 & \textbf{0.05} & 0.23 & 0.43 \\
    $6$ & Latent TransFuser~\cite{chitta2022transfuser} & 45.20 & 66.31 & 0.72 & 1.11 & 0.02 & 0.02 & \textbf{0.05} & 0.16 & 1.82 \\
    $7$ & GRIAD~\cite{chekroun2021gri} & 36.79 & 61.85 & 0.60 & 2.77 & \textbf{0.00} & 0.41 & 0.48 & 1.39 & 0.84 \\
    $8$ & TransFuser+~\cite{jaeger2021master} & 34.58 & 69.84 & 0.56 & 0.70 & 0.04 & 0.03 & 0.75 & 0.18 & 2.41 \\
    $9$ & Rails~\cite{chen2021learning} & 31.37 & 57.65 & 0.56 & 1.35 & 0.61 & 1.02 & 0.79 & 0.96 & 0.47 \\
    $10$ & IARL~\cite{toromanoff2020end} & 24.98 & 46.97 & 0.52 & 2.33 & \textbf{0.00} & 2.47 & 0.55 & 1.82 & 0.94 \\
    $11$ & NEAT~\cite{chitta2021neat} & 21.83 & 41.71 & 0.65 & 0.74 & 0.04 & 0.62 & 0.70 & 2.68 & 5.22 \\

    \bottomrule
\end{tabular}}
\caption{Comparison of our method and the state-of-the-art on the public CARLA leaderboard~\cite{leaderboard} (accessed Nov 2022). Methods are ranked by the driving score as the main metric. Driving Score, Route Completion, Infraction Score are higher the better, and the other metrics are lower the better. We outperform all other methods by a wide margin. We also lead the vehicle collision, offroad infraction numbers among all the methods.
}
\label{table:sota_detailed}
\end{table*}

\begin{table*}[t]
\centering
\scalebox{0.85}{
\begin{tabular}{c | c c | c c }
\hline

    ~ & \multicolumn{2}{c|}{\textbf{Town05 Short}} & \multicolumn{2}{c}{\textbf{Town05 Long}} \\
    \hline
    \textbf{Method} & Driving Score $\uparrow$ & Road Completion $\uparrow$ & Driving Score $\uparrow$ & Road Completion $\uparrow$ \\
    \hline
    CILRS~\cite{codevilla2019exploring} & 7.47$\pm$2.51 & 13.40$\pm$1.09 & 3.68$\pm$2.16 & 7.19$\pm$2.95 \\
    LBC~\cite{chen2020learning} & 30.97$\pm$4.17 & 55.01$\pm$5.14 & 7.05$\pm$2.13 & 32.09$\pm$7.40 \\
    TransFuser~\cite{prakash2021multi} & 54.52$\pm$4.29 & 78.41$\pm$3.75 & 33.15$\pm$4.04 & 56.36$\pm$7.14 \\
    NEAT~\cite{chitta2021neat}& 58.70$\pm$4.11 & 77.32$\pm$4.91 & 37.72$\pm$3.55 & 62.13$\pm$4.66 \\
    Roach~\cite{zhang2021end}& 65.26$\pm$3.63 & 88.24$\pm$5.16 & 43.64$\pm$3.95 & 80.37$\pm$5.68 \\
    WOR~\cite{chen2021learning}& 64.79$\pm$5.53 & 87.47$\pm$4.68 & 44.80$\pm$3.69 & 82.41$\pm$5.01 \\
    InterFuser~\cite{shao2022safety} & 94.95$\pm$1.91 & 95.19$\pm$2.57 & 68.31$\pm$1.86 & 94.97$\pm$2.87 \\
    \hline
    ReasonNet (Ours) & \textbf{95.71}$\pm$1.88 & \textbf{96.23}$\pm$3.17 & \textbf{73.22}$\pm$1.91 & \textbf{95.88}$\pm$2.31 \\
    
    \hline

\end{tabular}}

\caption{Comparison of our ReasonNet with six state-of-the-art methods in Town05 benchmark.  Our method outperformed other strong methods in all metrics and scenarios.}

\label{appendix: sota town05}
\end{table*}

\begin{table*}[t]
    \centering
    \scalebox{0.85}{
        \begin{tabular}{c |c c c}
        \hline
        \textbf{Method} &Driving Score $\uparrow$& Road Completion $\uparrow$& Infraction Score $\uparrow$\\
        \hline
        CILRS~\cite{codevilla2019exploring}  & 22.97$\pm$0.90 & 35.46$\pm$0.41 & 0.66$\pm$0.02\\
        LBC~\cite{chen2020learning} & 29.07$\pm$0.67 & 61.35$\pm$2.26 & 0.57$\pm$0.02  \\
        AIM~\cite{prakash2021multi}  & 51.25$\pm$0.17 & 70.04$\pm$2.31 & 0.73$\pm$0.03  \\
        TransFuser~\cite{prakash2021multi} & 53.40$\pm$4.54 &72.18$\pm$4.17 & 0.74$\pm$0.04 \\
        NEAT~\cite{chitta2021neat}  & 65.17$\pm$1.75 &79.17$\pm$3.25 & 0.82$\pm$0.01 \\
        Roach~\cite{zhang2021end}  & 65.08$\pm$0.99 & 85.16$\pm$4.20 & 0.77$\pm$0.02 \\
        WOR~\cite{chen2021learning} & 67.64$\pm$1.26  &90.16$\pm$3.81 & 0.75$\pm$0.02 \\
        InterFuser~\cite{shao2022safety} & 91.84$\pm$2.17 &  \textbf{97.12$\pm$1.95} & 0.95$\pm$0.02  \\
        \hline
        ReasonNet (Ours) & \textbf{93.25$\pm$2.91} &  96.84$\pm$2.17 & \textbf{0.96$\pm$0.02}  \\ \hline
    \end{tabular}}
    \caption{Comparison of our ReasonNet with other methods in CARLA 42 routes benchmark. Our method outperformed other strong methods in driving score and infraction score.}
    \label{appendix: sota 42 routes}
\end{table*}

\section{Implementation Details}
\label{appendix: Implementation Details}

\noindent\textbf{Model Details} The feature dimension of all decoders in our framework is set as 256. We use $K_e=1$, $K_{bev}=3$, $K_{opy}=3$, $C^{K}=64$, $C^{V}=256$ for the feature dimensions mentioned in Sec~\ref{sec: method}. The feature of the 5th stage in Resnet was used as the feature map $f_{i}$ in the 2D backbone. We use Fully Connected Layer and Batch Normalization \cite{ioffe2015batch} to construct a simplified version of PointNet\cite{qi2018frustum} to encode the information of raw LiDAR points in the 3D backbone.

\begin{figure}[h]
    \centering
    \includegraphics[width=0.45\textwidth]{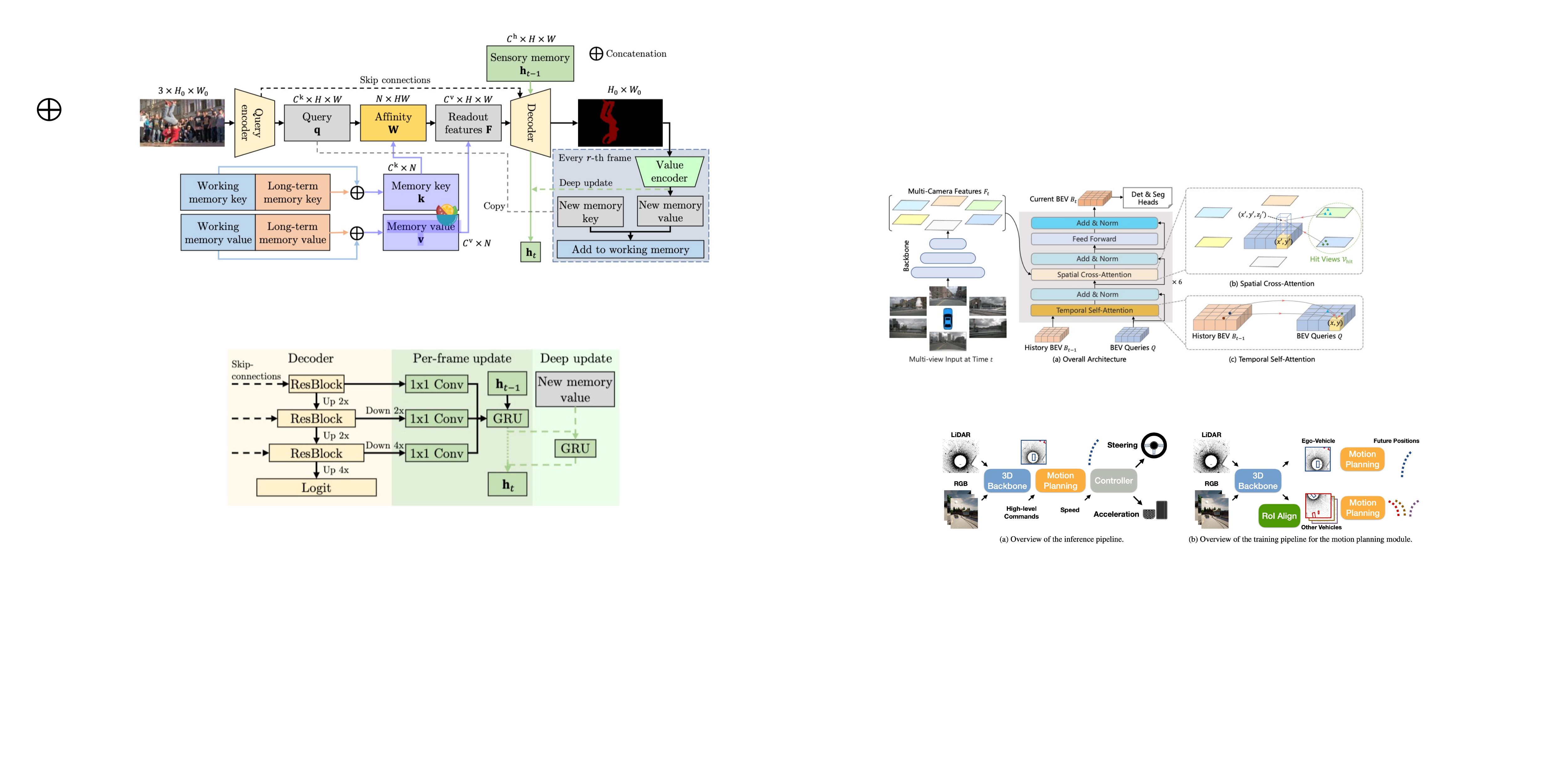}
    \caption{Overview of our pipeline for pretraining the perception module in the first training stage.}
    \label{fig:pipeline01}
\end{figure}

\noindent\textbf{Training} We train our models using the AdamW optimizer~\cite{loshchilov2018decoupled} and a cosine learning rate scheduler~\cite{loshchilov2016sgdr}. In the first training stage, the initial learning rate is set to $5e^{-4} \times \frac{Batch Size}{512}$ for the transformer encoder and the 3D backbone, and $2e^{-4} \times \frac{Batch Size}{512}$ for the 2D backbones. The weight decay is 0.07. We train the models for 35 epochs with the first 5 epochs for warm-up~\cite{he2016deep}. We used random scaling from $0.9\  \text{to}\ 1.1$ and color jittering to augment the collected RGB images. The overview of the first-stage framework can be found in Figure~\ref{fig:pipeline01}. In the second training stage, we freeze the perception module. The training schedule of the other two modules are similar to that in the first stage.

\noindent\textbf{Sensors} The RGB images are collected and cropped from one front-facing camera, two side-facing cameras, and one back-facing camera with a resolution of $800 \times 600$. Each camera has a 100$^\circ$ horizontal field of view (FOV), and the side cameras are angled at 60$^\circ$. For the front image, we scale the shorter side of the front camera input to 256 and crop its center patch of $224 \times 224$. For the focusing-view image, we directly crop the center of the front camera input to get a $128 \times 128$ patch. For the other images, the shorter side of the camera input is scaled to 160 and a center patch of $128 \times 128$ is taken.  

\noindent\textbf{Other hyper-parameter values}
Some other hyper-parameter values used in ReasonNet are listed in Table~\ref{table:parameter}.

\section{Benchmark details}
\label{appendix: Benchmark details}

We evaluate our method on the CARLA public leaderboard~\cite{leaderboard}, Town05 benchmark~\cite{prakash2021multi}, and our proposed DOS benchmark. Adversarial events\footnote{Adversarial events include unexpected agents rushing into the road from occluded regions, vehicles running red traffic lights, etc. Please refer to
https://leaderboard.carla.org/scenarios/ for detailed descriptions.} are included in the first two benchmarks, and occlusion events are included in the last benchmark. In these benchmarks, the ego vehicle is required to complete a given route without collision or traffic rules violation. 

\noindent\textbf{CARLA Leaderboard} The CARLA Autonomous Driving Leaderboard~\cite{leaderboard} is to evaluate the driving proficiency of autonomous agents in realistic traffic situations with a variety of weather conditions. The CARLA leaderboard provides a set of 76 routes for training and verifying agents and contains a secret set of 100 routes to evaluate the driving performance of the submitted agents.

\noindent\textbf{Town05 benchmark} In this benchmark, we use Town05 for evaluation and other towns for training. Following~\cite{prakash2021multi}, the benchmark includes two evaluation settings: 1) Town05 Short: 10 short routes of 100-500m, each comprising 3 intersections, 2) Town05 Long: 10 long routes of 1000-2000m, each comprising 10 intersections. Town05 is a complex town with multi-lane roads, single-lane roads, bridges, highways and exits. The core challenge of the benchmark is how to handle dynamic dense agents and adversarial events.

\noindent\textbf{CARLA 42 routes benchmark} The CARLA 42 routes benchmark was proposed in NEAT~\cite{chitta2021neat}, including six towns covering a variety of areas such as US-style intersections, EU-style intersections, freeways, roundabouts, stop signs, urban scenes and residential districts. The traffic density of each town is set to be comparable to busy traffic setting. We take the same configuration open-sourced by ~\cite{prakash2021multi} when we evaluated the methods.

\section{More Experimental results}
In this section we report additional experimental results, including the CARLA leaderboard and two other benchmarks.

\subsection{CARLA leaderboard}
Table~\ref{table:sota_detailed} shows the detailed comparison between our method and the baselines on the CARLA public Leaderboard~\cite{leaderboard}.
Our method also leads the vehicle collision and offroad infraction numbers among all the methods.

\begin{figure}[t]
    \centering
    \includegraphics[width=0.48\textwidth]{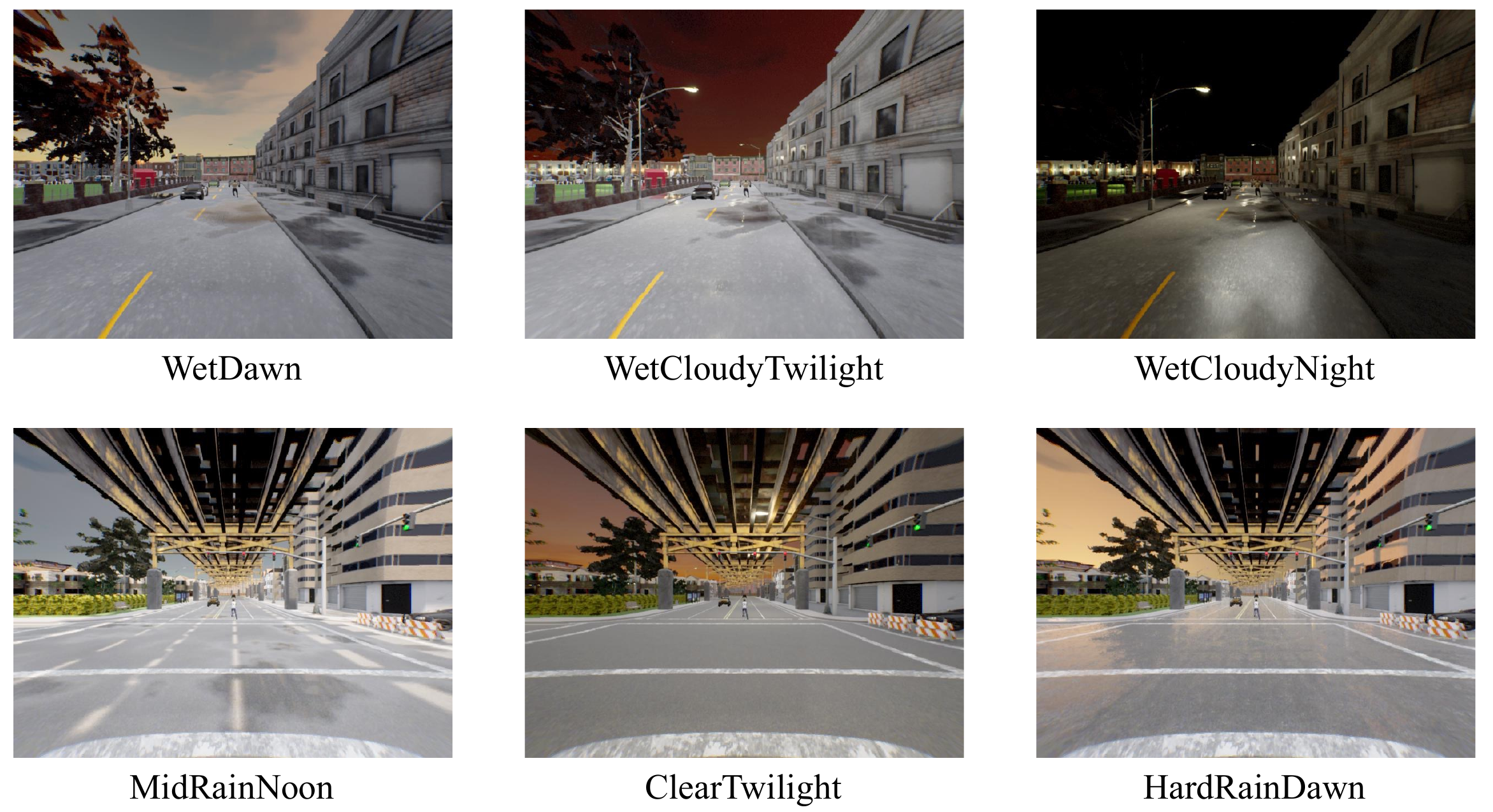}
    \caption{Different types of weather in our dataset.}
    \label{fig:weathers}
\end{figure}

\begin{table*}
\setlength{\tabcolsep}{10pt}
\centering
\resizebox{1.6\columnwidth}{!}{%
\begin{tabular}{llc}
\toprule
\textbf{Notation} & \textbf{Description} & \textbf{Value}  \\ 
\cmidrule(lr){1-3}
\multicolumn{3}{c}{BEV Map and Controller} \\
\cmidrule(lr){1-3}
$a_{max}$ & Maximum acceleration  & 1.0 m/s  \\
$v_{max}$ & Maximum velocity  & 7.5 m/$s^2$ \\
H, W & Size of the BEV map & 50, 50 \\
& Size of the BEV area & 50 meter $\times$ 50 meter \\
$H_{b}$ &  The detection range for the backward of the ego vehicle & 20 \\
& Scale factor for bounding box size of pedestrians and bicycles & 2 \\
\cmidrule(lr){1-3}

 \cmidrule(lr){1-3}
\multicolumn{3}{c}{Learning Process} \\
\cmidrule(lr){1-3}
 & Number of epochs & 35 \\
 & Number of warm-up epochs & 5 \\
$\lambda_{sign}$ & Weight for the traffic sign loss & 0.2 \\
$\lambda_{w}$ & Weight for the waypoints loss& 0.4 \\
$\lambda_{BEV}$ & Weight for the BEV map loss & 0.4 \\
$\lambda_{opy}$ & Weight for the occupancy map loss & 0.2 \\
$\lambda_{consistency}$ & Weight for the consistency loss & 0.05 \\
& Max norm for gradient clipping & 10.0 \\
& Weight decay & 0.07 \\
& Batch size & 256 \\
\bottomrule
\end{tabular}
}
\caption{The parameter used for ReasonNet.}
\label{table:parameter}
\end{table*}

\subsection{Town05 and CARLA 42 routes}
Table~\ref{appendix: sota town05} and Table~\ref{appendix: sota 42 routes} additionally compare the driving score, road completion, and infraction score of the presented approach to prior state-of-the-art on the CARLA Town05 benchmark~\cite{prakash2021multi} and CARLA 42 routes benchmark~\cite{chitta2021neat}.

\section{Data statistics}
We describe the detailed statistics for each town and their corresponding maps in Table~\ref{table:town_detail}. In Figure~\ref{fig:weathers}, we show six types of weathers among our dataset. For the submission for the online leaderboard, the model is trained in all eight towns. For the ablation studies, we train the models on five towns (Town01, Town03, Town04, Town06 ,and Town07).

\begin{table*}[h]
\centering
\scalebox{0.8}{
\begin{tabular}{@{}ccl@{}}
\toprule
Town Name & \#Frames & Description                             \\ \midrule
Town01    & 342846                                & A basic town layout consisting of ``T junctions"            \\
Town02    & 197240                                  & Similar to Town01, but smaller         \\
Town03    & 469115                                   & The most complex town, with a 5-lane junction, a roundabout, unevenness, a tunnel, and more                          \\
Town04    & 429979                                    & An infinite loop with a highway and a small town         \\
Town05    & 297140                                   & Squared-grid town with cross junctions and a bridge. It has multiple lanes per direction.  \\
Town06    & 148495                                 & Long highways with many highway entrances and exits. It also has a Michigan left                        \\
Town07    & 55299                               & A rural environment with narrow roads, barns and hardly any traffic lights    \\
Town10    & 69039                              & A city environment with different environments such as an avenue or promenade        \\ \bottomrule
\end{tabular}}
\caption{Detailed statistics of the number of frames and a brief description of each town.}
\label{table:town_detail}
\end{table*}

\section{License of Assets}
We use the open-source CARLA driving simulator~\cite{dosovitskiy2017carla}.
CARLA is released under the MIT license. Its assets are under the CC-BY license.
The pretrained ResNet model is under the MIT license.
The source code for our work will be publicly available once accepted and they are under the CC-BY-NC 4.0 license.

\end{document}